\documentclass[twoside,11pt]{article}

\usepackage{jmlr2e}

\jmlrheading{12}{2012}{??-??}{12/11}{??/??}{Philipp Hennig and
  Christian J. Schuler}

\ShortHeadings{Entropy Search}{Hennig and Schuler}
\title{Entropy Search\\ for Information-Efficient Global Optimization}
\author{\name Philipp Hennig \email phennig@tuebingen.mpg.de\\
\name Christian J. Schuler \email cschuler@tuebingen.mpg.de\\
\addr Department of Empirical Inference\\ 
Max Planck Institute for Intelligent Systems\\
Spemannstra\ss e 38, T\"ubingen, Germany}

\editor{}

\usepackage[cmex10]{amsmath}
\usepackage{amssymb}

\renewcommand{\Re}{\mathbb{R}}
\renewcommand{\L}{\mathcal{L}}
\newcommand{\N}{\mathcal{N}}

\newcommand{\Trans}{^{\intercal}}

\newcommand{\g}{\,|\,}

\renewcommand{\d}{\:\mathrm{d}}
\newcommand{\de}{\partial}

\newcommand{\vx}{\vec{x}}
\newcommand{\vy}{\vec{y}}

\newcommand{\vY}{\vec{Y}}
\newcommand{\vX}{\vec{X}}

\newcommand{\xmin}{x_{\min}}
\newcommand{\pmin}{p_{\min}}
\newcommand{\fmin}{f_{\min}}
\newcommand{\hpmin}{\hat{p}_{\min}}
\newcommand{\hqmin}{\hat{q}_{\min}}

\renewcommand{\vec}{\boldsymbol}

\renewcommand{\O}{\mathcal{O}}

\newcommand{\Id}{\vec{I}}

\newcommand{\argmin}{\operatorname*{arg\: min}}
\newcommand{\chol}{\operatorname{\mathsf{C}}}

\newcommand{\Exp}{\mathsf{E}}
\newcommand{\KL}{\text{KL}}

\usepackage{tikz}
\usetikzlibrary{arrows,shapes,plotmarks}

\tikzset{>=stealth'} 
\tikzstyle{graphnode} = 
   [circle,draw=black,minimum size=22pt,text centered,text
     width=22pt,inner sep=0pt] 
\tikzstyle{obs} = [graphnode,fill=black,text=white]
\tikzstyle{var} = [graphnode,fill=white,text=black]
\tikzstyle{fac} = [rectangle,draw=black,fill=black!25,minimum size=5pt]
\tikzstyle{edge}= [draw=white,double=black,thick,-]

\usepackage{pgfplots}
\usetikzlibrary{arrows,shapes,plotmarks}

\DeclareSymbolFont{stmry}{U}{stmry}{m}{n}
\DeclareMathSymbol\leftrightarrowtriangle\mathrel{stmry}{"5D}
\DeclareMathSymbol\leftarrowtriangle\mathrel{stmry}{"5E}
\DeclareMathSymbol\rightarrowtriangle\mathrel{stmry}{"5F}
\DeclareMathSymbol\mapstochar\mathrel{stmry}{"5A}

\renewcommand{\gets}{\leftarrowtriangle}
\renewcommand{\to}{\rightarrowtriangle}

\usepackage{algorithm}
\usepackage{algpseudocode}
\algrenewcommand{\algorithmiccomment}[1]{{\footnotesize \hfill$\rhd$ #1}}

\newif\iffinal 

\pgfrealjobname{EntropySearch}  
\iffinal
 
\else
 
\fi

\begin{document}

\maketitle

\begin{abstract}
  Contemporary global optimization algorithms are based on local
  measures of utility, rather than a probability measure over location
  and value of the optimum. They thus attempt to collect low function
  values, not to learn about the optimum. The reason for the absence
  of probabilistic global optimizers is that the corresponding
  inference problem is intractable in several ways. This paper
  develops desiderata for probabilistic optimization algorithms, then
  presents a concrete algorithm which addresses each of the
  computational intractabilities with a sequence of approximations and
  explicitly adresses the decision problem of maximizing information
  gain from each evaluation.
\end{abstract}


\section{Introduction}
\label{sec:introduction}

Optimization problems are ubiquitous in science, engineering, and
economics. Over time the requirements of many separate fields have led
to a heterogenous set of settings and algorithms. Speaking very
broadly, however, there are two distinct regimes for optimization. In
the first one, relatively cheap function evaluations take place on a
numerical machine and the goal is to find a ``good'' region of low or
high function values. Noise tends to be small or negligible, and
derivative observations are often available at low additional cost;
but the parameter space may be very high-dimensional. This is the
regime of \emph{numerical, local} or \emph{convex} optimization, often
encountered as a sub-problem of machine learning algorithms. Popular
algorithms for such settings include quasi-Newton methods
\citep{broyden1965class,fletcher1970new,goldfarb1970family,shanno1970conditioning},
the conjugate gradient method \citep{hestenes1952methods}, and
stochastic optimization and evolutionary search methods (e.g.\
\citet{hansen2001completely}), to name only a few. Since these
algorithms perform local search, constraints on the solution space are
often a crucial part of the problem. Thorough introductions can be
found in the textbooks by \citet{nocedal1999numerical} and
\citet{boyd2004convex}. This paper will utilize algorithms from this
domain, but it is not its primary subject.

In the second milieu, which this paper addresses, the function itself
is not known and needs to be learned during the search for its
\emph{global} minimum within some measurable (usually: bounded)
domain. Here, the parameter space is often relatively low-dimensional,
but evaluating the function involves a monetarily or morally expensive
physical process -- building a prototype, drilling a borehole, killing
a rodent, treating a patient. Noise is often a nontrivial issue, and
derivative observations, while potentially available, cannot be
expected in general. While algorithms for such applications need to be
tractable, their most important desideratum is efficient use of data,
rather than raw computational cost. This domain is often called
\emph{global optimization}, but is also closely associated with the
field of \emph{experimental design} and related to the concept of
\emph{exploration} in reinforcement learning. The learned model of the
function is also known as a \emph{response surface} in some
communities. The two contributions of this paper are a probabilistic
view on this field, and a concrete algorithm for such problems.

\subsection{Problem Definition}
\label{sec:problem-definition}
We define the problem of \emph{probabilistic global optimization}: Let
$I\subset\Re^D$ be some bounded domain of the real vector space. There
is a function $f:I\to\Re$, and our knowledge about $f$ is described by
a probability measure $p(f)$ over the space of functions
$I\to\Re$. This induces a measure
\begin{equation}
  \label{eq:1}
  \pmin(x) \equiv p[x=\argmin f(x)] = \int_{f:I\to\Re} p(f)
  \prod_{\stackrel{\tilde{x}\in I}{\tilde{x} \neq
      x}} \theta[f(\tilde{x})-f(x)] \d f
\end{equation}
were $\theta$ is Heaviside's step function. The exact meaning of the
``infinite product'' over the entire domain $I$ in this equation
should be intuitively clear, but is defined properly in the
Appendix. Note that the integral is over the infinite-dimensional
space of functions. We assume we can evaluate the function\footnote{We
  may further consider observations of linear operations on $f$. This
  includes derivative and integral observations of any order, if they
  exist. Section \ref{sec:deriv-observ} addresses this point; it is
  unproblematic under our chosen prior, but clutters the notation, and
  is thus left out elsewhere in the paper.} at any point $x\in I$
within some bounded domain $I$, obtaining function values $y(x)$
corrupted by noise, as described by a likelihood $p(y\g
f(x))$. Finally, let $L(x^*,\xmin)$ be a loss function describing the
cost of naming $x^*$ as the result of optimization if the true minimum
is at $\xmin$. This loss function induces a loss functional
$\L(\pmin)$ assigning utility to the uncertain knowledge about
$\xmin$, as
\begin{equation}
  \label{eq:3}
  \L(\pmin) = \int_I [\min_{x^*} L(x^*,\xmin)] \pmin(\xmin) \d \xmin.
\end{equation}
The goal of global optimization is to decrease the expected loss after
$H$ function evaluations at locations $\vx=\{x_1,\dots,x_H\}\subset
I$. The expected loss is
\begin{equation}
  \label{eq:2}
  \langle \L \rangle_H = \int p(\vy \g \vx) \L(\pmin(x \g \vy,\vx))
  \d \vy =  \iint p(\vy \g \vec{f}(\vx)) p(\vec{f}(\vx)\g\vx) \L(\pmin(x \g \vy,\vx))
  \d \vy \d \vec{f}
\end{equation}
where $\L(\pmin(x \g \vy,\vx))$ should be understood as the cost
assigned to the measure $\pmin(x)$ induced by the posterior belief
over $f$ after observations $\vy=\{y_1,\dots,y_H\}\subset\Re$ at the
locations $\vx$.

The remainder of this paper will replace the symbolic objects in this
general definition with concrete measures and models to construct an
algorithm we call \emph{Entropy Search}. But it is useful to pause at
this point to contrast this definition with other concepts of
optimization.

\paragraph{Probabilistic Optimization}
\label{sec:prob-optim}
The distinctive aspect of our definition of ``optimization'' is
Equation \eqref{eq:1}, an explicit role for the function's
extremum. Previous work did not consider the extremum so directly. In
fact, many frameworks do not even use a measure over the function
itself. An example of optimizers that only implicitly encode
assumptions about the function are genetic algorithms
\citep{schmitt2004theory} and evolutionary search
\citep{hansen2001completely}. If such formulations feature the global
minimum $\xmin$ at all, then only in statements about the limit
behavior of the algorithm after many evaluations. Not explicitly
writing out the prior over the function space can have advantages:
Probabilistic analyses tend to involve intractable integrals; a less
explicit formulation thus allows to construct algorithms with
interesting properties that would be entirely intractable from a
probabilistic viewpoint. But non-probabilistic algorithms cannot make
explicit statements about the location of the minimum. At best, they
may be able to provide bounds.

Fundamentally, reasoning about optimization of functions on continuous
domains \emph{after finitely many evaluations}, like any other
inference task on spaces without natural measures, is impossible
without prior assumptions. For intuition, consider the following
thought experiment: Let $(\vx_0,\vy_0)$ be a finite, possibly empty, set
of previously collected data. For simplicity, and without loss of
generality, assume there was no measurement noise, so the true
function actually passes through each data point. Say we want to
suggest that the minimum of $f$ may be at $x^* \in I$. To make this
argument, we propose a number of functions that pass through
$(\vx_0,\vy_0)$ and are minimized at $x^*$. We may even suggest an
uncountably infinite set of such functions. Whatever our proposal, a
critic can always suggest another uncountable set of functions that
also pass through the data, and are \emph{not} minimized at $x^*$. To
argue with this person, we need to reason about the relative size of
our set versus their set. Assigning size to infinite sets amounts to
the aforementioned normalized measure over admissible functions
$p(f)$, and the consistent way to reason with such measures is
probability theory \citep{kolmogorov_axioms,cox1946probability}. Of
course, this amounts to imposing assumptions on $f$, but this is a
fundamental epistemological limitation of inference, not a special
aspect of optimization.

\paragraph{Relationship to the Bandit Setting}
\label{sec:diff-band-sett}

There is a considerable amount of prior work on continuous bandit
problems, also sometimes called ``global optimization''
\citep[e.g.][]{kleinberg2005nearly,grunewalder2010regret,srinivasgaussian}. The
bandit concept differs from the setting defined above, and bandit
regret bounds do not apply here: Bandit algorithms seek to minimize
\emph{regret}, the sum over function values at evaluation points,
while probabilistic optimizers seek to infer the minimum, no matter
what the function values at evaluation points. An optimizer gets to
evaluate $H$ times, then has to make one single decision regarding
$\L(\pmin)$. Bandit players have to make $H$ evaluations, such that
the evaluations produce low values. This forces bandits to focus their
evaluation policy on function value, rather than the loss at the
horizon (see also Section \ref{sec:model-comparison}). In
probabilistic optimization, the only quantity that counts is the
quality of the belief on $\pmin$ under $\L$, after $H$ evaluations,
not the sum of the function values returned during those $H$
steps.

\paragraph{Relationship to Heuristic Gaussian Process Optimization and
Response Surface Optimization}
\label{sec:relat-heur-gauss}

There are also a number of works employing Gaussian process measures
to construct heuristics for search, also known as ``Gaussian process
global optimization''
\citep{jones1998efficient,lizotte2008practical,osborne2009gaussian}.
As in our definition, these methods explicitly infer the function from
observations, constructing a Gaussian process posterior. But they then
evaluate at the location maximizing a heuristic $u[p(f(x))]$ that
turns the \emph{marginal} belief over $f(x)$ at $x$, which is a
univariate Gaussian $p(f(x))=\N[f(x);\mu(x),\sigma^2(x)]$, into an ad
hoc utility for evaluation, designed to have high value at locations
close to the function's minimum. Two popular heuristics are the
\emph{probability of improvement} \citep{lizotte2008practical}
\begin{equation}
  \label{eq:15}
  u_{\text{PI}}(x) = p[f(x)<\eta] = \int_{-\infty} ^\eta
  \N(f(x);\mu(x),\sigma(x)^2) \d f(x) = \Phi\left( \frac{\eta-\mu(x)}{\sigma(x)} \right)
\end{equation}
and \emph{expected improvement} \citep{jones1998efficient}
\begin{equation}
  \label{eq:16}
  u_{\text{EI}}(x) = \Exp[\min\{0,(\eta-f(x))\}] = (\eta - \mu)
  \Phi\left( \frac{\eta-\mu(x)}{\sigma(x)} \right) + \sigma \phi\left( \frac{\eta-\mu(x)}{\sigma(x)} \right)
\end{equation}
where $\Phi(z)=1/2[1+\operatorname{erf}(z/\sqrt{2})]$ is the standard
Gaussian cumulative density function, $\phi(x)=\N(x;0,1)$ is the
standard Gaussian probability density function, and $\eta$ is a
current ``best guess'' for a low function value, e.g.\ the lowest
evaluation so far. 

These two heuristics have different units of measure: probability of
improvement is a probability, expected improvement has the units of
$f$. Both utilities differ markedly from Eq.\ \eqref{eq:1}, $\pmin$,
which is a probability \emph{measure} and as such a \emph{global}
quantity. See Figure \ref{fig:samples} for a comparison of the three
concepts on an example. The advantage of the heuristic approach is
that it is computationally lightweight, because the utilities have
analytic form. But local measures cannot capture general decision
problems of the type described above. For example, these algorithms do
not capture the effect of evaluations on knowledge: A small region of
high density $\pmin(x)$ may be less interesting to explore than a
broad region of lower density, because the expected \emph{change} in
knowledge from an evaluation in the broader region may be much larger,
and may thus have much stronger effect on the loss. If the goal is to
infer the location of the minimum (more generally: minimize loss at
the horizon), the optimal strategy is to evaluate where we expect to
\emph{learn} most about the minimum (reduce loss toward the horizon),
rather then where we think the minimum \emph{is} (recall Section
\ref{sec:diff-band-sett}). The former is a nonlocal problem, because
evaluations affect the belief, in general, everywhere. The latter is a
local problem.

\section{Entropy Search}
\label{sec:concrete-algorithm}

The probable reason for the absence of global optimization algorithms
from the literature is a number of intractabilities in any concrete
realisation of the setting of Section
\ref{sec:problem-definition}. This section makes some choices and
constructs a series of approximations, to arrive at a tangible
algorithm, which we call \emph{Entropy Search}. The derivations evolve
along the following path.
\begin{description}
\item[choosing $p(f)$] We commit to a Gaussian process prior on $f$
  (Section \ref{sec:gauss-proc-meas}). Limitations and implications of
  this choice are outlined, and possible extensions suggested, in
  Sections \ref{sec:deriv-observ} and \ref{sec:limit-extens-gauss}.
\item[discretizing $\pmin$] We discretize the problem of calculating
  $\pmin$, to a finite set of representer points chosen from a
  non-uniform measure, which deals gracefully with the curse of
  dimensionality. Artefacts created by this discretization are studied
  in the tractable one-dimensional setting (Section
  \ref{sec:discr-repr-cont}).
\item[approximating $\pmin$] We construct an efficient approximation
  to $\pmin$, which is required because Eq.\ \eqref{eq:1}, even for
  finite-dimensional Gaussian measures, is not analytically tractable,
  (Section \ref{sec:appr-pmin-with}). We compare the approximation to
  the (asymptotically exact, but more expensive) Monte Carlo solution.
\item[predicting change to $\pmin$] The Gaussian process measure affords
  a straightforward but rarely used analytic probabilistic formulation
  for the \emph{change} of $p(f)$ as a function of the next evaluation
  point (Section \ref{sec:pred-innov-from}).
\item[choosing loss function] We commit to relative \emph{entropy}
  from a uniform distribution as the loss function, as this can be
  interpreted as a utility on gained \emph{information} about the
  location of the minimum (Section \ref{sec:information-gain-log}).
\item[predicting expected information gain] From the predicted change,
  we construct a first-order expansion on $\langle \L \rangle$ from
  future evaluations and, again, compare to the asymptotically exact
  Monte Carlo answer (Section \ref{sec:first-order-appr}).
\item[choosing greedily] Faced with the exponential cost of the exact
  dynamic problem to the horizon $H$, we accept a greedy approach for
  the reduction of $\langle \L\rangle$ at every step. We illustrate
  the effect of this shortcut in an example setting (Section
  \ref{sec:defect-greedy-plann}).
\end{description}

\subsection{Gaussian Process Measure on $f$}
\label{sec:gauss-proc-meas}

\begin{figure}
  \centering
   \beginpgfgraphicnamed{figures/GPonly-external}%
   \input{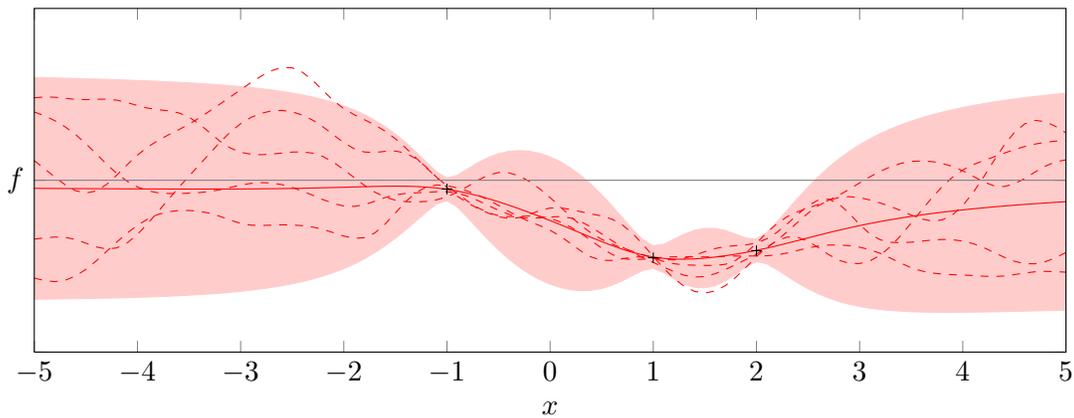}%
   \endpgfgraphicnamed%
 
  \caption{A Gaussian process measure (rational quadratic kernel),
    conditioned on three previous observations (black crosses). Mean
    function in solid red, marginal standard deviation at each
    location (two standard deviations) as light red tube. Five sampled
    functions from the current belief as dashed red lines. Arbitrary
    ordinate scale, zero in gray.}
  \label{fig:GP_init}
\end{figure}

The remainder of the paper commits to Gaussian process measures for
$p(f)$. These are convenient for the task at hand due to their
descriptive generality and their convenient analytic properties. Since
this paper is aimed at readers from several communities, this section
contains a very brief introduction to some relevant aspects of
Gaussian processes; readers familiar with the subject can safely skip
ahead. A thorough introduction can be found in a textbook of
\citet{RasmussenWilliams}. Some readers from other fields may find it
helpful to know that more or less special cases of Gaussian process
inference are elsewhere known under names like \emph{Kriging}
\citep{krigestatistical} and \emph{Kolmogorov-Wiener prediction}
\citep{wiener1957prediction}, but while these frameworks are
essentially the same idea, the generality of their definitions varies,
so restrictions of those frameworks should not be assumed to carry
over to Gaussian process inference as understood in machine learning.

A Gaussian process is an infinite-dimensional probability density,
such that each linear finite-dimensional restriction is multivariate
Gaussian. The infinite-dimensional space can be thought of as a space
of functions, and the finite-dimensional restrictions as \emph{values}
of those functions at locations $\{x^* _i \}_{i=1,\dots,N}$. Gaussian
process beliefs are parametrized by a \emph{mean function} $m:I\to\Re$
and a \emph{covariance function} $k:I \times I \to \Re$. For our
particular analysis, we restrict the domain $I$ to finite, compact
subsets of the real vector spaces $\Re^D$. The covariance function,
also known as the \emph{kernel}, has to be positive definite, in the
sense that any finite-dimensional matrix with elements $K_{ij} =
k(x_i,x_j)$ has to be positive definite $\forall x_i,x_j\in I$. A
number of such kernel functions are known in the literature, and
different kernel functions induce different kinds of Gaussian process
measures over the space of functions. Among the most widely used
kernels for regression are the \emph{squared exponential} kernel
\begin{equation}
  \label{eq:5}
  k_\text{SE}(x,x';\vec{S},s) = s^2 \exp\left[-\frac{1}{2} (x - x'
    )\Trans \vec{S}^{-1}(x-x')  \right]
\end{equation}
which induces a measure that puts nonzero mass on only smooth
functions of \emph{characteristic length-scale} $\vec{S}$ and
\emph{signal variance} $s^2$ \citep{mackay1998introduction}, and the
\emph{rational quadratic} kernel
\citep{matérn1960spatial,RasmussenWilliams}
\begin{equation}
  \label{eq:6}
  k_{\text{RQ}}(x,x';\vec{S},s,\alpha) = s^2 \left(1 + \frac{1}{2\alpha} (x - x'
    )\Trans \vec{S}^{-1}(x-x')\right)^{-\alpha}
\end{equation}
which induces a belief over smooth functions whose characteristic
length scales are a scale mixture over a distribution of width
$1/\alpha$ and location $\vec{S}$. Other kernels can be used to induce
beliefs over non-smooth functions \citep{matérn1960spatial}, and even
over non-continuous functions \citep{uhlenbeck1930theory}. Experiments
in this paper use the two kernels defined above, but the results apply
to all kernels inducing beliefs over \emph{continuous}
functions. While there is a straightforward relationship between
kernel continuity and the mean square continuity of the induced
\emph{process}, the relationship between the kernel function and the
continuity of each \emph{sample} is considerably more involved
\citep[][\textsection3]{adler1981geometry}. Regularity of the kernel
also plays a nontrivial role in the question wether the distribution
of infima of samples from the process is well-defined at all
\citep{AdlerSuprema}. In this work, we side-step this issue by
assuming that the chosen kernel is sufficiently regular to induce a
well-defined belief $\pmin$ as defined by Equation \eqref{eq:14}.

Kernels form a semiring: products and sums of kernels are kernels.
These operations can be used to generalize the induced beliefs over
the function space (Section \ref{sec:limit-extens-gauss}). Without loss
of generality, the mean function is often set to $m\equiv 0$ in
theoretical analyses, and this paper will keep with this tradition,
except for Section \ref{sec:limit-extens-gauss}. Where $m$ is nonzero,
its effect is a straightforward off-set $p(f(x)) \to p(f(x)-m(x))$.

For the purpose of regression, the most important aspect of Gaussian
process priors is that they are conjugate to the likelihood from
finitely many observations $(\vX,\vY)=\{\vx_i,y_i\}_{i=1,\dots,N}$ of
the form $y_i(\vx_i)=f(\vx_i)+\xi$ with Gaussian noise $\xi\sim
\N(0,\sigma^2)$. The posterior is a Gaussian process with mean and
covariance functions
\begin{equation}
  \label{eq:7}
  \mu(\vx^*) = k_{\vx^*,\vX} [K_{\vX,\vX} + \sigma^{2}\Id]^{-1} \vy \quad;
  \quad \Sigma(\vx^*,\vx_*) = k_{\vx^*,\vx_*} - k_{\vx^*,\vX} [K_{\vX,\vX} +
  \sigma^{2}\Id]^{-1} k_{\vX,\vx_*}
\end{equation}
where $K_{\vX,\vX}$ is the kernel Gram matrix $K_{\vX,\vX}
^{(i,j)}=k(x_i,x_j)$, and other objects of the form
$k_{\vec{a},\vec{b}}$ are also matrices with elements
$k_{\vec{a},\vec{b}} ^{(i,j)} = k(\vec{a}_i,\vec{b}_j)$. Finally, for
what follows it is important to know that it is straightforward to
sample ``functions'' (point-sets of arbitrary size from $I$) from a
Gaussian process. To sample the value of a particular sample at the
$M$ locations $\vX^*$, evaluate mean and variance function as a
function of any previously collected datapoints, using Eq.\
\eqref{eq:7}, draw a vector $\vec{\zeta}\sim\prod ^M \N(0,1)$ of $M$
random numbers i.i.d.\ from a standard one-dimensional Gaussian
distribution, then evaluate
\begin{equation}
  \label{eq:8}
  \tilde{f}(\vX^*) = \mu(\vX^*) + \chol[\Sigma(\vX^*,\vX^*)]\Trans \vec{\zeta}
\end{equation}
where the operator $\chol$ denotes the Cholesky decomposition
\citep{Cholesky}.

\subsection{Discrete Representations for Continuous Distributions}
\label{sec:discr-repr-cont}
\begin{figure}
  \centering
   \beginpgfgraphicnamed{figures/pmin-external}%
   \input{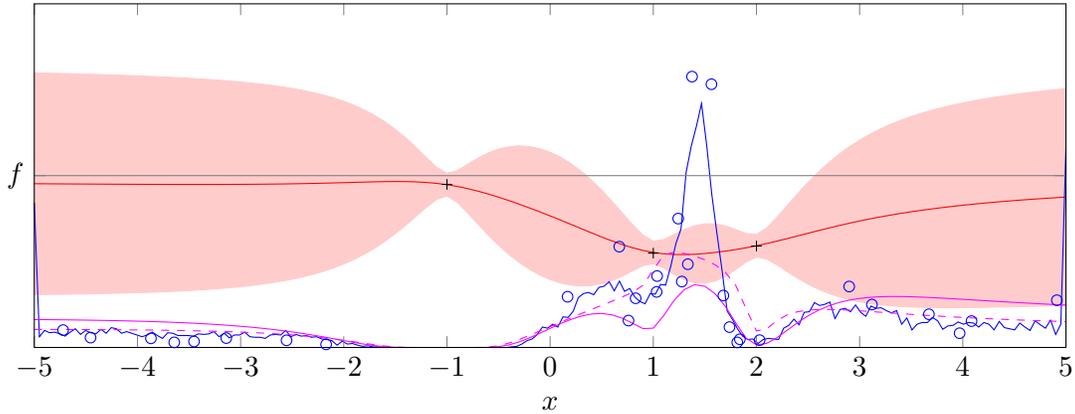}%
   \endpgfgraphicnamed%
 
  \caption{$\pmin$ induced by $p(f)$ from Figure \ref{fig:GP_init}.
    $p(f)$ repeated for reference. Blue solid line: Asymptotically
    exact representation gained from exact sampling of functions on a
    regular grid. For comparison, the plot also shows the local
    utilities \emph{probability of improvement} (dashed magenta) and
    \emph{expected improvement} (solid magenta) often used for
    Gaussian process global optimization. Blue circles: Approximate
    representation on representer points, sampled from probability of
    improvement measure. Stochastic error on sampled values, due to
    only asymptotically correct assignment of mass to samples, and
    varying density of points, focusing on relevant areas of
    $\pmin$. This plot uses arbitrary scales for each object: The two
    heuristics have different units of measure, differing from that of
    $\pmin$. Notice the interesting features of $\pmin$ at the
    boundaries of the domain: The prior belief encodes that $f$ is
    smooth, and puts finite probability mass on the hypothesis that
    $f$ has negative (positive) derivative at the right (left)
    boundary of the domain. With nonzero probability, the minimum thus
    lies exactly on the boundary of the domain, rather than within a
    Taylor radius of it.}
  \label{fig:samples}
\end{figure}

Having established a probability measure $p(f)$ on the function, we
turn to constructing the belief $\pmin(x)$ over its
minimum. Inspecting Equation \eqref{eq:1}, it becomes apparent that it
is challenging in two ways: First, because it is an integral over an
infinite-dimensional space, and second, because even on a
finite-dimensional space it may be a hard integral for a particular
$p(f)$. This section deals with the former issue, the following
Section \ref{sec:appr-pmin-with} with the latter.

It may seem daunting that $\pmin$ involves an infinite-dimensional
integral. The crucial observation for a meaningful approximation in
finite time is that regular functions can be represented meaningfully
on finitely many points. If the stochastic process representing the
belief over $f$ is sufficiently regular, then Equation \eqref{eq:1}
can be approximated arbitrarily well with finitely many representer
points. The discretization grid need not be regular -- it may be
sampled from any distribution which puts non-zero measure on every
open neighborhood of $I$. This latter point is central to a graceful
handling of the curse of dimensionality: The na\"ive approach of
approximately solving Equation \eqref{eq:1} on a regular grid, in a
$D$-dimensional domain, would require $\O(\exp(D))$ points to achieve
any given resolution. This is obviously not efficient: Just like in
other numerical quadrature problems, any given resolution can be
achieved with fewer representer points if they are chosen irregularly,
with higher resolution in regions of greater influence on the result
of integration. We thus choose to \emph{sample} representer points
from a proposal measure $u$, using a Markov chain Monte Carlo sampler
(our implementation uses shrinking rank slice sampling
\citep{thompson2010slice}).

What is the effect of this stochastic discretization? A non-uniform
quadrature measure $u(\tilde{x})$ for $N$ representer locations
$\{\tilde{x}_i\}_{i=1,\dots,N}$ leads to varying widths in the
``steps'' of the representing staircase function. As $N\to\infty$, the
width of each step is approximately proportional to $(u(\tilde{x}_i)
N)^{-1}$. Section \ref{sec:appr-pmin-with} will construct a
discretized $\hqmin(\tilde{x}_i)$ that is an approximation to the
probability that $\fmin$ occurs within the step at $\tilde{x}_i$. So
the approximate $\hpmin$ on this step is proportional to
$\hqmin(\tilde{x}_i) u(\tilde{x}_i)$, and can be easily normalized
numerically, to become an approximation to $\pmin$.

How should the measure $u$ be chosen? Unfortunately, the result of the
integration, being a density rather than a function, is itself a
function of $u$, and the loss-function is also part of the
problem. So it is nontrivial to construct an optimal quadrature
measure. Intuitively, a good proposal measure for discretization
points should put high resolution on regions of $I$ where the shape of
$\pmin$ has strong influence on the loss, and on its change. For our
choice of loss function (Section \ref{sec:information-gain-log}), it
is a good idea to choose $u$ such that it puts high mass on regions of
high value for $\pmin$. But for other functions, this need not always
be the case.

We have experimented with a number of ad-hoc choices for $u$, and
found the aforementioned ``expected improvement'' and ``probability of
improvement'' (Section \ref{sec:relat-heur-gauss}) to lead to
reasonably good performance. We use these functions for a similar
reason as their original authors: Because they \emph{tend} to have
high value in regions where $\pmin$ is also large. To avoid confusion,
however, note that we use these functions as unnormalized measures to
\emph{sample discretization points} for our \emph{calculation} of
$\pmin$, not as an approximation for $\pmin$ itself, as was done in
previous work by other authors. Defects in these heuristics have
weaker effect on our algorithm than in the cited works: In our case,
if $u$ is not a good proposal measure, we simply need more samples to
construct a good representation of $\pmin$. In the limit of
$N\to\infty$, all choices of $u$ perform equally well, as long as they
put nonzero mass on all open neighborhoods of the domain.

\subsection{Approximating $\pmin$ with Expectation Propagation}
\label{sec:appr-pmin-with}
\begin{figure}
  \centering
  \begin{tikzpicture}[node distance=2cm]
    \node[var] at (0,0) (f) {$\vec{f}$};
    \node[fac, left of=f] (pf) {} edge (f);
    \node[fac, right of=f] (ff) {} edge (f);
    \node[anchor=east] at (pf.west)
    {$\N(\vec{f};\tilde{\vec\mu},\tilde{\vec\Sigma})$};
    \node[anchor=west] at (ff.east) 
    {$\theta[f(\tilde{x}_j) - f(\tilde{x}_i)]$};
    \draw (1.0,-1) rectangle (5.5,1);
    \node[anchor=south east] at (5.5,-1) {$i\neq j$};
  \end{tikzpicture}
  \caption{Graphical model providing motivation for EP approximation
    on $\pmin$. See text for details.}
  \label{fig:EP}
\end{figure}
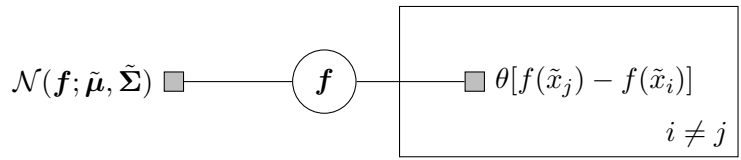

\begin{figure}
  \centering
   \beginpgfgraphicnamed{figures/EP_pmin-external}%
   \input{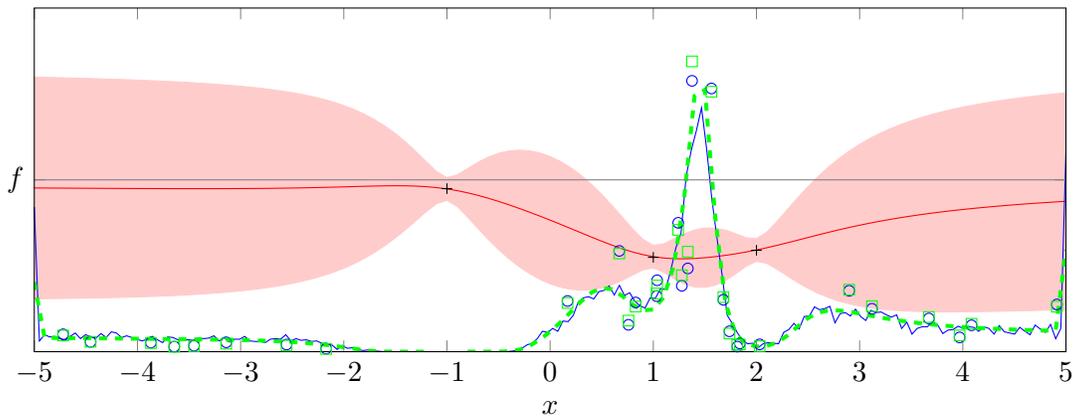}%
   \endpgfgraphicnamed%
 
  \caption{EP-approximation to $\pmin$ (green). Other plots as in
    previous figures. EP achieves good agreement with the
    asymptotically exact Monte Carlo approximation to $\pmin$,
    including the point masses at the boundaries of the domain.}
  \label{fig:EP_toMC}
\end{figure}

The previous Section \ref{sec:discr-repr-cont} provided a way to
construct a non-uniform grid of $N$ discrete locations $\tilde{x}_i$,
$i=1,\dots,N$. The restriction of the Gaussian process belief to these
locations is a multivariate Gaussian density with mean
$\tilde{\vec{\mu}}\in\Re^N$ and covariance
$\tilde{\vec{\Sigma}}\in\Re^{N\times N}$. So Equation
\eqref{eq:1} reduces to a discrete probability \emph{distribution} (as
opposed to a density)
\begin{equation}
  \label{eq:17}
  \hpmin(x_i) = \int_{\vec{f}\in\Re^N} \N(\vec{f};\tilde{\vec{\mu}},\tilde{\vec{\Sigma}})
  \prod_{i\neq j} ^N \theta(f(x_j)-f(x_i)) \d \vec{f}.
\end{equation}
This is a multivariate Gaussian integral over a half-open, convex,
piecewise linearly constrained integration region -- a polyhedral
cone. Unfortunately, such integrals are known to be intractable
\citep{plackett1954reduction,Holly_NInt}. However, it is possible to
construct an effective approximation $\hqmin$ based on Expectation
Propagation (EP) \citep{EP_Minka}: Consider the belief
$p(f(\tilde{x}))$ as a ``prior message'' on $f(\tilde{x})$, and each
of the terms in the product as one factor providing another
message. This gives the graphical model shown in Figure
\ref{fig:EP}. Running EP on this graph provides an approximate
Gaussian marginal, whose normalisation constant $\hqmin(x_i)$, which
EP also provides, approximates $p(f\g \xmin=x_i)$. The EP algorithm
itself is somewhat involved, and there are a number of algorithmic
technicalities to take into account for this particular setting. We
refer interested readers to recent work by \citet{EPMGP}, which gives
a detailed description of these aspects. The cited work also
establishes that, while EP's approximations to Gaussian integrals are
not always reliable, in this particular case, where there are as many
constraints as dimensions to the problem, the approximation is
generally of high quality (see Figure \ref{fig:EP_toMC} for an
example). An important advantage of the EP approximation over both
numerical integration and Monte Carlo integration (see next Section)
is that it allows analytic differentiation of $\hqmin$ with respect to
the parameters $\tilde{\vec\mu}$ and $\tilde{\vec\Sigma}$
\citep{EPMGP,SeegerEP}. This fact will become important in Section
\ref{sec:first-order-appr}.

The computational cost of this approximation is considerable: Each
computation of $\hqmin(\tilde{x}_i)$, for a given $i$, involves $N$
factor updates, which each have rank 1 and thus cost $\O(N^2)$. So,
overall, the cost of calculating $\hqmin(\tilde\vx)$ is
$\O(N^4)$. This means $N$ is effectively limited to well below
$N=1000$. Our implementation uses a default of $N=50$, and can
calculate next evaluation points in $\sim 10$ seconds. Once again, it
is clear that this algorithm is not suitable for simple numerical
optimization problems; but a few seconds are arguably an acceptable
waiting time for physical optimization problems.

\subsubsection{An alternative: Sampling}
\label{sec:an-altern-sampl}
An alternative to EP is Monte Carlo integration: sample $S$ functions
exactly from the Gaussian belief on $p(f)$, at cost $O(N^2)$ per
sample, then find the minimum for each sample in $\O(N)$ time. This
technique was used to generate the asymptotically exact plots in
Figures \ref{fig:samples} and following. It has overall cost
$\O(SN^3)$, and can be implemented efficiently using Matrix-Matrix
multiplications, so each evaluation of this algorithm is considerably
faster than EP. It also has the advantage of asymptotic
exactness. But, unfortunately, it provides no analytic derivatives,
because of strong discontinuity in the step functions of Eq.\
\eqref{eq:1}. So the choice is between a first-order expansion using
EP (see Section \ref{sec:first-order-appr}) which is expensive, but
provides a re-usable, differentiable function, and repeated calls to a
cheaper, asymptotically exact sampler. In our experiments, the former
option appeared to be considerably faster, and of acceptable
approximative quality. But for relatively high-dimensional
optimization problems, where one would expect to require relatively
large $N$ for acceptable discretization, the sampling approach can be
expected to scale better. The code we plan to publish upon acceptance
offers a choice between these two approaches.

\subsection{Predicting Innovation from Future Observations}
\label{sec:pred-innov-from}
\begin{figure}
  \centering
   \beginpgfgraphicnamed{figures/Innovation-external}%
   \input{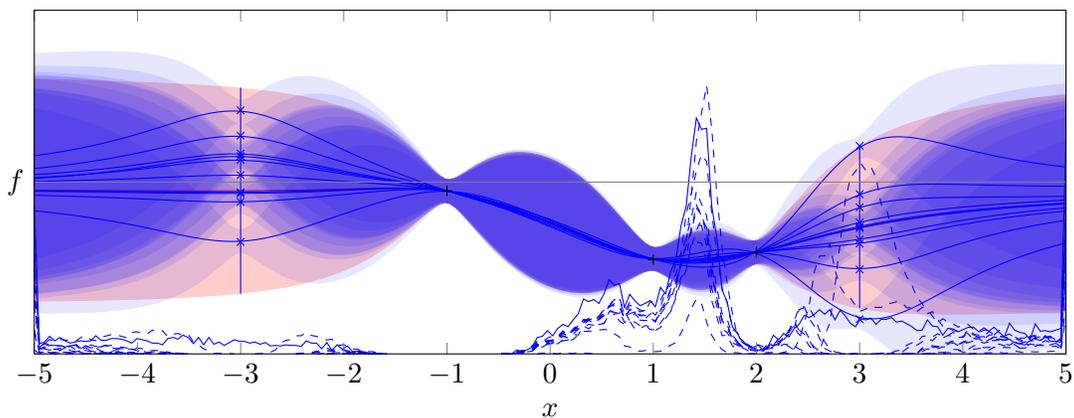}%
   \endpgfgraphicnamed%
 
  \caption{ Innovation from two observations at $x=-3$ and
    $x=3$. Current belief in red, as in Figure
    \ref{fig:GP_init}. Samples from the belief over possible beliefs
    after observations at $\vx$ in blue. For each sampled innovation,
    the plot also shows the induced innovated $\pmin$ (lower sampling
    resolution as previous plots). Innovations from several (here:
    two) observations can be sampled jointly.}
  \label{fig:innovation}
\end{figure}

As detailed in Equation \eqref{eq:2}, the optimal choice of the next
$H$ evaluations is such that the \emph{expected} change in the loss
$\langle \L \rangle_{\vx}$ is minimal, i.e.\ effects the biggest
possible expected drop in loss. The loss is a function of $\pmin$,
which in turn is a function of $p(f)$. So predicting change in loss
requires predicting change in $p(f)$ as a function of the next
evaluation points. It is another convenient aspect of Gaussian
processes that they allow such predictions in analytic form
\citep{GaussianRL}: Let previous observations at $\vX_0$ have yielded
observations $\vY_0$. Evaluating at locations $\vX$ will give new
observations $\vY$, and the mean will be given by
\begin{equation}
  \label{eq:18}
  \begin{aligned}
    \mu(x^*) &= [k_{x^*,\vX_0}, k_{x^*,\vX}]
    \begin{pmatrix}
      K_{\vX_0,\vX_0} & k_{\vX_0,\vX} \\ k_{\vX,\vX_0} & K_{\vX,\vX}
    \end{pmatrix}^{-1}
    \begin{pmatrix}
      \vY_0 \\ \vY
    \end{pmatrix} \\
    &= k_{x^*,\vX_0}K^{-1} _{\vX_0,\vX_0} \vY_0 + (k_{x^*,\vX} -
    k_{x^*,\vX_0} K^{-1} _{\vX_0,\vX_0} k_{\vX_0,\vX}) \times \\ &
    \qquad (k_{\vX,\vX} - k_{\vX,\vX_0} K^{-1} _{\vX_0,\vX_0}
    k_{\vX_0,\vX})^{-1}
    (\vY - k_{\vX,\vX_0} K^{-1} _{\vX_0,\vX_0} \vY_0)\\
    & = \mu_0(x^*) + \Sigma_0(x^*,\vX) \Sigma^{-1} _0 (\vX,\vX) (\vY -
    \mu_0(\vX))
 \end{aligned}
\end{equation}
where $K_{a,b} ^{(i,j)} = k(a_i,b_j) + \delta_{ij} \sigma^2$. The step
from the first to the second line involves an application of the
matrix inversion lemma, the last line uses the mean and covariance
functions conditioned on the dataset $(\vX_0,\vY_0)$ so far. Since
$\vY$ is presumed to come from this very Gaussian process belief, we
can write
\begin{equation}
  \label{eq:20}
\vY = \mu(\vX) +
\chol[\Sigma(\vX,\vX)]\Trans \vec{\Omega}' + \sigma\vec{\omega} = \mu(\vX) +
\chol[\Sigma(\vX,\vX) + \sigma^2 \Id_H]\Trans\vec\Omega \qquad
\vec\Omega,\vec\Omega',\vec{\omega} \sim \N(0,\Id_H),
\end{equation}
and Equation \eqref{eq:18} simplifies. An even simpler construction
can be made for the covariance function. We find that mean and
covariance function of the posterior after observations $(\vX,\vY)$
are mean and covariance function of the prior, incremented by the
\emph{innovations}
\begin{equation}
  \label{eq:19}
  \begin{aligned}
    \Delta\mu_{\vX,\Omega}(x^*) &= \Sigma(x^*,\vX)
    \Sigma^{-1}(\vX,\vX)
    \chol[\Sigma(\vX,\vX)+\sigma^2\Id_H] \vec{\Omega} \\
    \Delta\Sigma_{\vX}(x^*,x_*) &= \Sigma(x^*,\vX)
    \Sigma^{-1}(\vX,\vX) \Sigma(\vX,x_*).
  \end{aligned}
\end{equation}
The change to the mean function is stochastic, while the change to the
covariance function is deterministic. Both innovations are functions
both of $\vX$ and of the evaluation points $x^*$. One use of this
result is to sample $\langle \L\rangle_{\vX}$ by sampling innovations,
then evaluating the innovated $\pmin$ for each innovation in an inner
loop, as described in Section \ref{sec:an-altern-sampl}. An
alternative, described in the next section, is to construct an
analytic first order approximation to $\langle \L\rangle_{\vX}$ from
the EP prediction constructed in Section \ref{sec:appr-pmin-with}. As
mentioned above, the advantage of this latter option is that it
provides an analytic function, with derivatives, which allows
efficient numerical local optimization.

\subsection{Information Gain -- the Log Loss}
\label{sec:information-gain-log}

\begin{figure}
  \centering
   \beginpgfgraphicnamed{figures/LogLoss-external}%
   \input{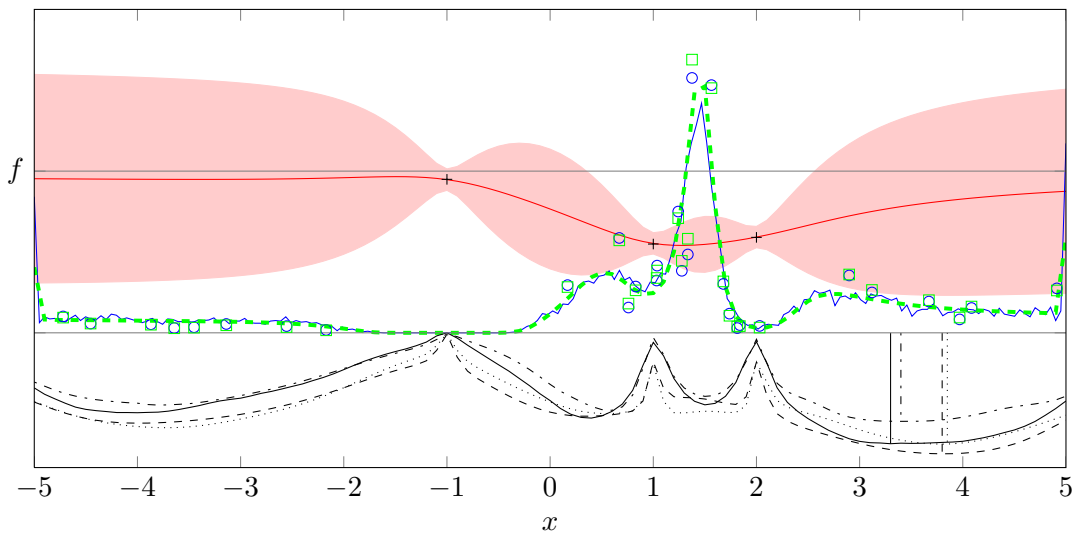}%
   \endpgfgraphicnamed%
 
  \caption{1-step predicted loss improvement for the log loss
    (relative entropy). Upper part of plot as before, for
    reference. Monte Carlo prediction on regular grid as solid black
    line. Monte Carlo prediction from sampled irregular grid as
    dot-dashed black line. EP prediction on regular grid as black
    dashed line. EP prediction from samples as black dashed line. The
    minima of these functions, where the algorithm will evaluate next,
    are marked by vertical lines. While the predictions from the
    various approximations are not identical, they lead to similar
    next evaluation points. Note that these next evaluation points
    differ qualitatively from the choice of the GP optimization
    heuristics of Figure \ref{fig:samples}. Since each approximation
    is only tractable up a multiplicative constant, the scales of
    these plots are arbitrary, and only chosen to overlap for
    convenience.}
  \label{fig:logloss}
\end{figure}

To solve the decision problem of where to evaluate the function next
in order to learn most about the location of the minimum, we need to
say what it means to ``learn''. Thus, we require a loss functional
that evaluates the information content of innovated beliefs $\pmin$.
This is, of course, a core idea in information theory. The seminal
paper by \citet{shannon1948mathematical} showed that the negative
expectation of probability logarithms,
\begin{equation}
  \label{eq:24}
  \mathsf{H}[\vec{p}] = -\langle \log p \rangle_{\vec p} = -\sum_i p_i \log p_i
\end{equation}
known as entropy, has a number of properties that allow its
interpretation as a measure of uncertainty represented by a
probability distribution $p$. It's value can be be interpreted as the
number of natural information units an optimal compression algorithm
requires to encode a sample from the distribution, given knowledge of
the distribution. However, it has since been pointed out repeatedly
that this concept does not trivially generalize to probability
densities. A density $p(x)$ has a unit of measure $[x]^{-1}$, so its
logarithm is not well-defined, and one cannot simply replace summation
with integration in Equation \eqref{eq:24}. A functional that
\emph{is} well-defined on probability densities and preserves many of
the information-content interpretations of entropy
\citep{jaynes2003probability} is \emph{relative entropy}, also known
as Kullback-Leibler divergence \citep{kullback1951information}. We use
its negative value as a loss function emphasizing information gain.
\begin{equation}
  \label{eq:25}
  \L_{\KL}(p; b) = -\int p(x) \log\frac{p(x)}{b(x)} \d x
\end{equation}
As base measure $b$ we choose the uniform measure $\mathbb{U}_I
(x)=|I|^{-1}$ over $I$, which is well-defined because $I$ is presumed
to be bounded\footnote{Although uniform measures appeal as a
  natural representation of ignorance, they do encode an assumption
  about $I$ being represented in a ``natural'' way. Under a nonlinear
  transformation of $I$, the distribution would not remain
  uniform. For example, uniform measures on the [0,1] simplex appear
  bell-shaped in the softmax basis \citep{mackay1998choice}. So, while
  $b$ here does not represent prior knowledge on $\xmin$ per se, it
  does provide a unit of measure to information and as such is
  nontrivial.}. With this choice, the loss is maximized (at $\L=0$)
for a uniform belief over the minimum, and diverges toward negative
infinity if $p$ approaches a Dirac point distribution. The resulting
algorithm, Entropy Search, will thus choose evaluation points such
that it expects to move away from the uniform base measure toward a
Dirac distribution as quickly as possible.

The reader may wonder: What about the alternative idea of maximizing,
at each evaluation, entropy relative to the \emph{current} $\pmin$?
This would only encourage the algorithm to attempt to change the
current belief, but not necessarily in the right direction. For
example, if the current belief puts very low mass on a certain region,
an evaluation that has even a small chance of increasing $\pmin$ in
this region could appear more favorable than an alternative evaluation
predicted to have a large effect on regions where the current $\pmin$
has larger values. The point is not to just change $\pmin$, but to
change it such that it moves away from the base measure.

Recall that we approximate the \emph{density} $p(x)$ using a
\emph{distribution} $\hat{p}(x_i)$ on a finite set $\{x_i\}$ of
representer points, which define steps of width proportional, up to
stochastic error, to an unnormalized measure $\tilde{u}(x_i)$. In
other words, we can approximate $\pmin(x)$ as
\begin{equation}
  \label{eq:26}
  \pmin(x) \approx \frac{\hat{p}(x_i) N \tilde{u}(x_i)}{Z_u} \qquad Z_u =
  \int \tilde{u}(x) \d x \qquad x_i = \argmin_{\{x_j\}} \| x - x_j \|.
\end{equation}
We also note that after $N$ samples, the unit element of measure has
size, up to stochastic error, of $\Delta x_i \approx
\frac{Z_u}{\tilde{u}(x_i) N}$. So we can approximately represent the
loss
\begin{equation}
  \label{eq:27}
  \begin{aligned}
    \L_{\KL}(\pmin;b) &\approx -\sum_i \pmin(x_i) \Delta x_i
    \log\frac{p_{\min}(x_i)}{b(x_i)} \\
    &= -\sum_i \hat{p}_{\min}(x_i) \log
    \frac{\hat{p}_{\min}(x_i) N \tilde{u}(x_i)}{Z_u b(x_i)} \\
    &= -\sum_i \hat{p}_{\min}(x_i) \log \frac{\hat{p}_{\min} (x_i)
      \tilde{u}(x_i)}{b(x_i)} + \log\left(\frac{Z_u}{N}\right) \sum_i \hat{p}_{\min} (x_i)\\
    &= \mathsf{H}[\hat{p}_{\min}] - \langle\log\tilde{u}\rangle_{\hat{p}_{\min}} +
    \langle \log b\rangle_{\hat{p}_{\min}} + \log Z_u - \log N
  \end{aligned}
\end{equation}
which means we do not require the normalization constant $Z_u$ for
optimization of $\L_\KL$. For our uniform base measure, the third term
in the last line is a constant, too; but other base measures would contribute
nontrivially.

\subsection{First-Order Approximation to $\langle \L\rangle $}
\label{sec:first-order-appr}

Since EP provides analytic derivatives of $\pmin$ with respect to mean
and covariance of the Gaussian measure over $f$, we can construct a
first order expansion of the expected change in loss from
evaluations. To do so, we consider, in turn, the effect of evaluations
at $\vX$ on the measure on $f$, the induced change in $\pmin$, and
finally the change in $\L$. Since the change to the mean is Gaussian
stochastic, It{\=o}'s Lemma \citep{ito1951stochastic} applies. The
following Equation uses the summation convention: double indices in
products are summed over.
\begin{multline}
  \label{eq:21}
  \langle \Delta\L \rangle_{\vX} = \int \L\bigg[\pmin ^0 + \frac{\de
    \pmin}{\de \Sigma(\tilde{x}_i,\tilde{x}_j)} \Delta\Sigma_{\vX}
  (\tilde{x}_i,\tilde{x}_j) + \frac{\de^2 \pmin}{\de \mu_i \de \mu_j}
  \Delta\mu_{\vX,\vec{1}}(\tilde{x}_i)\Delta\mu_{\vX,\vec{1}}(\tilde{x}_j)\\
  + \frac{\de \pmin}{\de \mu(\tilde{\vx}_i)}
  \Delta_{\vX,\Omega}\mu(\tilde{x}_i) +
  \O((\Delta\mu)^2,(\Delta\Sigma)^2) \bigg] \N(\Omega;0,1) \d \Omega -
  \L[\pmin^0].
\end{multline}
The first line contains deterministic effects, the first term in the
second line covers the stochastic aspect. Monte Carlo integration over
the stochastic effects can be performed approximately using a small
number of samples $\vec\Omega$. These samples should be drawn only
once, at first calculation, to get a differentiable function
$\langle\Delta\L\rangle_{\vX}$ that can be re-used in subsequent
optimization steps.

The above formulation is agnostic with respect to the loss
function. Hence, in principle, Entropy Search should be easy to
generalize to different loss functions. But recall that the fidelity
of the calculation of Equation \eqref{eq:21} depends on the
intermediate approximate steps, in particular the choice of
discretization measure $\tilde{u}$. We have experimented with other
loss functions and found it difficult to find a good measure
$\tilde{u}$ providing good performance for many such loss
functions. So this paper is limited to the specific choice of the
relative entropy loss function. Generalization to other losses is
future work.

\subsection{Greedy Planning, and its Defects}
\label{sec:defect-greedy-plann}

\begin{figure}
  \centering
   \beginpgfgraphicnamed{figures/dH2-external}%
%
%
\begin{tikzpicture}

\begin{axis}[%
scale only axis,
width=0.5\textwidth,
height=0.5\textwidth,
y dir=reverse,
xmin=-5.05, xmax=5.05,
ymin=-5.05, ymax=5.05,
axis on top]
\addplot graphics [xmin=-5.050000e+000, xmax=5.050000e+000, ymin=-5.050000e+000, ymax=5.050000e+000] {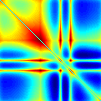};
\end{axis}
\end{tikzpicture}%
   \endpgfgraphicnamed%
 
  \caption{Expected drop in relative entropy (see Section
    \ref{sec:information-gain-log}) from two additional evaluations to
    the three old evaluations shown in previous plots. First new
    evaluation on abscissa, second new evaluation on ordinate, but due
    to the exchangeability of Gaussian process measures, the plot is
    symmetric. Diagonal elements excluded for numerical reasons. Blue
    regions are more beneficial than red ones. The relatively
    complicated structure of this plot illustrates the complexity of
    finding the optimal $H$-step evaluation locations.}
  \label{fig:dh2}
\end{figure}

The previous sections constructed a means to predict, approximately,
the expected drop in loss from $H$ new evaluations at locations
$\vX=\{\vx_i\}_{i=1,\dots,N}$. The remaining task is to optimize these
locations. It may seem pointless to construct an optimization
algorithm which itself contains an optimization problem, but note that
this new optimization problem is quite different from the initial
one. It is a numerical optimization problem, of the form described in
Section \ref{sec:introduction}: We can evaluate the utility function
numerically, without noise, with derivatives, and at hopefully
relatively low cost compared to the physical process we are ultimately
trying to optimize.

Nevertheless, one issue remains: Optimizing evaluations over the
entire horizon $H$ is a dynamical programming problem, which, in
general, has cost exponential in $H$. However, this problem has a
particular structure: Apart from the fact that evaluations drawn from
Gaussian process measures are exchangeable, there is also other
evidence that optimization problems are benign from the point of view
of planning. For example, \citet{srinivasgaussian} show that the
information gain over the function values is submodular, so that
greedy learning of the function comes close to optimal learning of the
function. While is is not immediately clear whether this statement
extends to our issue of learning about the function's minimum, it is
obvious that the greedy choice of whatever evaluation location most
reduces expected loss in the immediate next step is guaranteed to
never be catastrophically wrong. In contrast to general planning,
there are no ``dead ends'' in inference problems. At worst, a greedy
algorithm may choose an evaluation point revealed as redundant by a
later step. But thanks to the consistency of Bayesian inference in
general, and Gaussian process priors in particular
\citep{van2011information}, no decision can lead to an evaluation that
somehow makes it impossible to learn the true function afterward. In
our approximate algorithm, we thus adopt this greedy approach. It
remains an open question for future research whether approximate
planning techniques can be applied efficiently to improve performance
in this planning problem.

\subsection{Further Issues}
\label{sec:further-issues}
This section digresses from the main line of thought to briefly touch
upon some extensions and issues arising from the choices made in
previous sections. For the most part, we point out well-known analytic
properties and approximations that can be used to generalize the
algorithm. Since they apply to Gaussian process regression rather than
the optimizer itself, they will not play a role in the empirical
evaluation of Section \ref{sec:loss-functions}.

\subsubsection{Derivative Observations}
\label{sec:deriv-observ}
Gaussian process inference remains analytically tractable if instead
of, or in addition to direct observations of $f$, we observe the
result of any \emph{linear} operator acting on $f$. This includes
observations of the function's derivatives \citep[][\textsection
9.4]{RasmussenWilliams} and, with some caveats, to integral
observations \citep{minka2000deriving}. The extension is pleasingly
straightforward: The kernel defines covariances between function
values. Covariances between the function and its derivatives are
simply given by
\begin{equation}
  \label{eq:9}
  \operatorname{cov}\left(\frac{\de^n f(\vx)}{\prod_i \de x_i} , \frac{\de^m
      f(\vx')}{\prod_j \de x' _j}\right) = \frac{\de^{n+m}
    k(\vx,\vx')}{\prod_i \de x_i \prod_j \de x' _j}
\end{equation}
so kernel evaluations simply have to be replaced with derivatives (or
integrals) of the kernel where required. Obviously, this operation is
only valid as long as the derivatives and integrals in question exist
for the kernel in question. Hence, all results derived in previous
sections for optimization from function evaluations can trivially be
extended to optimization from function and derivative observations, or
from only derivative observations. 

\subsubsection{Learning Hyperparameters}
\label{sec:learn-hyperp}

Throughout this paper, we have assumed kernel and likelihood function
to be given. In real applications, this will not usually be the
case. In such situations, the hyperparameters defining these two
functions, and if necessary a mean function, can be learned from the
data, either by setting them to maximum likelihood values, or by
full-scale Bayesian inference using Markov chain Monte Carlo
methods. See \citet[][\textsection 5]{RasmussenWilliams} and
\citet{murray2010slice} for details. In the latter case, the belief
$p(f)$ over the function is a mixture of Gaussian processes. To still
be able to use the algorithm derived so far, we approximate this
belief with a single Gaussian process by calculating expected values
of mean and covariance function.

Ideally, one would want to take account of this hierarchical learning
process in the decision problem addressed by the optimizer. This adds
another layer of computation complexity to the problem, and is outside
of the scope of this paper. Here, we contend ourselves with
considering the uncertainty of the Gaussian process conditioned on a
particular set of hyperparameters. 

\subsubsection{Limitations and Extensions of Gaussian Processes for
  Optimization}
\label{sec:limit-extens-gauss}

\begin{figure}
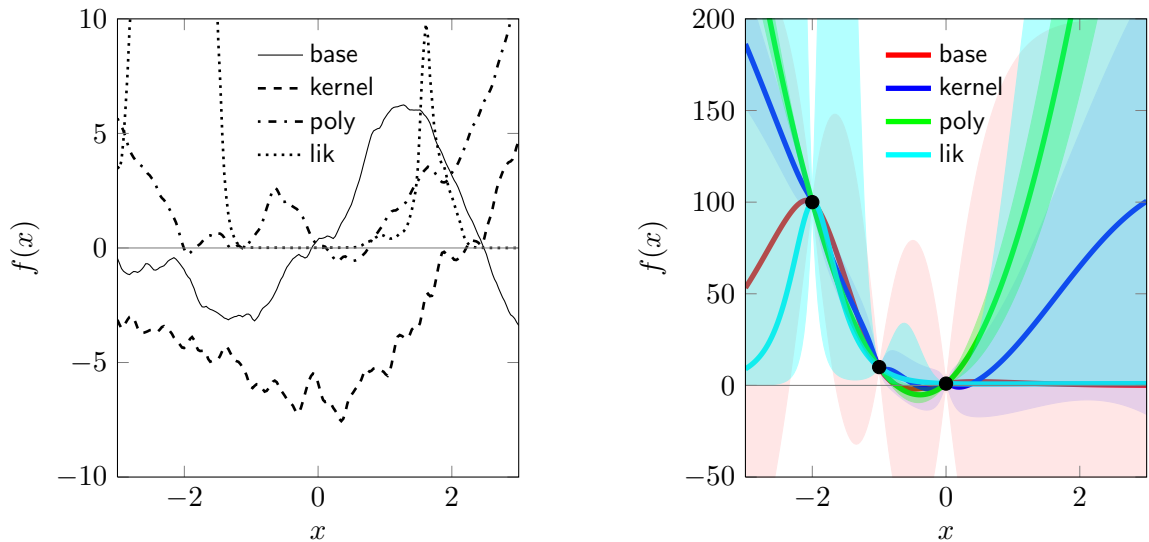

  \centering
   \beginpgfgraphicnamed{figures/priors-external}%
   \input{figures/priors.tikz}%
   \endpgfgraphicnamed%
  \hfill
   \beginpgfgraphicnamed{figures/posteriors-external}%
   \input{figures/posteriors.tikz}%
   \endpgfgraphicnamed%
 
  \caption{Generalizing GP regression. {\bfseries Left:} Samples from
    different priors. {\bfseries Right:} Posteriors (mean, two
    standard deviations) after observing three datapoints with
    negligible noise (kernel parameters differ between the two
    plots). {\sf base}: standard GP regression with Mat\'ern
    kernel. {\sf kernel}: sum of two kernels (square exponential and
    rational quadratic) of different length scales and strengths. {\sf
      poly}: polynomial (here: quadratic) mean function. {\sf lik}:
    Non-Gaussian likelihood (here: logarithmic link function). The
    scales of both $x$ and $f(x)$ are functions of kernel parameters,
    so the numerical values in this plot have relevance only relative
    to each other. Note the strong differences in both mean and
    covariance functions of the posteriors.}%
 \label{fig:kernel-extensions}
\end{figure}

Like any probability measure over functions, Gaussian process measures
are not arbitrarily general. In particular, the most widely used
kernels, including the two mentioned above, are \emph{stationary},
meaning they only depend on the difference between locations, not
their absolute values. Loosely speaking, the prior ``looks the same
everywhere''. One may argue that many real optimization problems do
not have this structure. For example, it may be known that the
function tends to have larger functions values toward the boundaries
of $I$ or, more vaguely, that it is roughly
``bowl-shaped''. Fortunately, a number of extensions readily suggest
themselves to address such issues (Figure
\ref{fig:kernel-extensions}).

\begin{description}
\item[Parametric Means] As pointed out in Section
  \ref{sec:gauss-proc-meas}, we are free to add any parametric general
  linear model as the mean function of the Gaussian process.
  \begin{equation}
    \label{eq:10}
    m(x) = \sum_i \phi_i(x) w_i
  \end{equation}
  Using Gaussian beliefs on the weights $w_i$ of this model, this
  model may be learned at the same time as the Gaussian process itself
  \citep[][\textsection 2.7]{RasmussenWilliams}. Polynomials such as
  the quadratic $\vec{\phi}(\vx) = [\vx;\vx\vx\Trans]$ are beguiling
  in this regard, but they create an explicit ``origin'' at the center
  of $I$, and induce strong long-range correlations between opposite
  ends of $I$. This seems pathological: In most settings, observing
  the function on one end of $I$ should not tell us much about the
  value at the opposite end of $I$. But we may more generally choose
  any feature set for the linear model. For example, a set of radial
  basis functions $\phi_i(\vx) = \exp(\|\vx-\vec{c}_i\|^2 / \ell_i
  ^2)$ around locations $\vec{c}_i$ at the rims of $I$ can explain
  large function values in a region of width $\ell_i$ around such a
  feature, without having to predict large values at the center of
  $I$. This idea can be extended to a nonparametric version, described
  in the next point.
\item[Composite Kernels] Since kernels form a semiring, we may sum a
  kernel of large length scale and large signal variance and a kernel
  of short length scale and low signal variance. For example
  \begin{equation}
    \label{eq:11}
    k(x,x') = k_{\text{SE}} (x,x';s_1,\vec{S}_1) +
    k_\text{RQ}(x,x',s_2,\vec{S}_2,\alpha_2) \qquad s_1 \gg s_2;
    \vec{S}_1 ^{ij} \gg \vec{S}_2 ^{ij} \;\forall i,j
  \end{equation}
  yields a kernel over functions that, within the bounded domain $I$,
  look like ``rough troughs'': global curvature paired with local
  stationary variations. A disadvantage of this prior is that it
  thinks ``domes'' just as likely as ``bowls''. An advantage
  is that it is a very flexible framework, and does not induce
  unwanted global correlations.
\item[Nonlinear Likelihoods] An altogether different effect can be
  achieved by a non-Gaussian, non-linear likelihood function. For
  example, if $f$ is known to be strictly positive, one may assume the
  noise model
  \begin{equation}
    \label{eq:12}
    p(y\g g) = \N(y; \exp(g), \sigma^2); \quad f = \exp(g)
  \end{equation}
  and learn $g$ instead of $f$. Since the logarithm is a convex
  function, the minimum of the latent $g$ is also a minium of $f$. Of
  course, this likelihood leads to a non-Gaussian posterior. To retain
  a tractable algorithm, approximate inference methods can be used to
  construct approximate Gaussian posteriors. In our example (labeled
  $\mathsf{lik}$ in Figure \ref{fig:kernel-extensions}), we used a
  Laplace approximation: It is straightforward to show that Equation
  \eqref{eq:12} implies
  \begin{equation}
    \label{eq:13}
    \left.\frac{\de \log p(y\g g)}{\de g}\right|_{g=\hat{g}}
  \operatorname*{=}^{!} 0 \quad \Rightarrow \hat{g}=\log y \qquad
\left.  \frac{\de^2 \log p(y\g g)}{\de^2 g}\right|_{g=\hat{g}} = \frac{y^2}{\sigma^2}
  \end{equation}
  so a Laplace approximation amounts to a heteroscedastic noise model,
  in which an observation $(y,\sigma^2)$ is incorporated into the
  Gaussian process as $(\log(y),(\sigma/y)^2)$. This approximation is
  valid if $\sigma \ll y$ (see Figure 3). For functions on logarithmic
  scales, however, finding minima smaller than the noise level, at
  logarithmic resolution, is a considerably harder problem anyway.
\end{description}

\begin{figure}%
\centering
   \beginpgfgraphicnamed{figures/logarithms-external}%
   \input{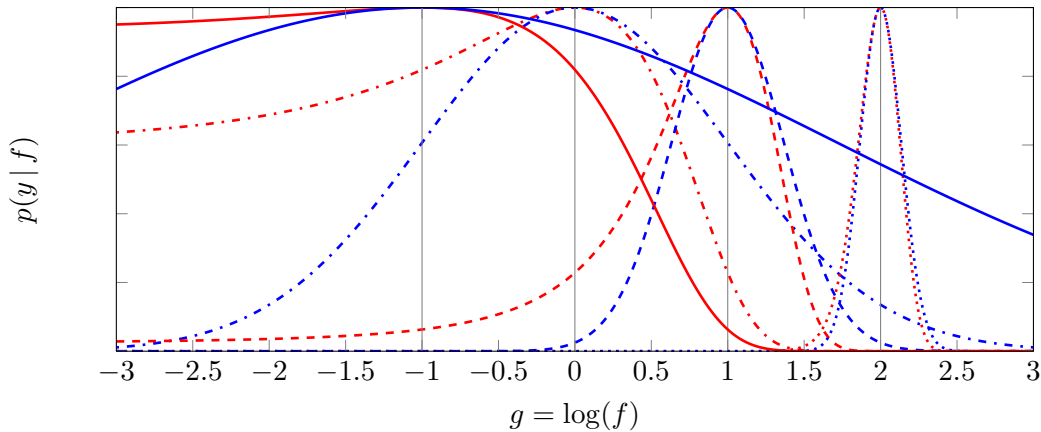}%
   \endpgfgraphicnamed%
 
\caption{Laplace approximation for a logarithmic Gaussian
  likelihood. True likelihood in red, Gaussian approximation in blue,
  maximum likelihood solution marked in grey. Four log relative values
  $a=\log(y/\sigma)$ of sample $y$ and noise $\sigma$ (scaled in
  height for readability). $a=-1$ (solid); $a=0$ (dash-dotted); $a=1$
  (dashed); $a=2$ (dotted). The approximation is good for $a\gg0$.}%
\label{fig:log}%
\end{figure}

The right part of Figure \ref{fig:kernel-extensions} shows posteriors
produced using the three approaches detailed above, and the base case
of a single kernel with strong signal variance, when presented with
the same three data points, with very low noise.  The strong
difference between the posteriors may be disappointing, but it is a
fundamental aspect of inference: Different prior assumptions lead to
different posteriors, and function space inference is impossible
without priors. Each of the four beliefs shown in the Figure may be
preferable over the others in particular situations. The polynomial
mean describes functions that are “almost parabolic”. The exponential
likelihood approximation is appropriate for functions with an
intrinsic logarithmic scale. The sum kernel approach is pertinent for
the search for local minima of globally stationary functions. Classic
methods based on polynomial approximations are a lot more restrictive
than any of the models described above.

Perhaps the most general option is to use additional prior information
$\mathcal{I}$ giving $p(\xmin\g\mathcal{I})$, independent of $p(f)$,
to encode outside information about the location of the
minimum. Unfortunately, this is intractable in general. But it may be
approached through approximations. This option is outside of the scope
of this paper, but will be the subject of future work.

\subsection{Summary -- the Entire Algorithm}
\label{sec:summ-entire-algor}

\begin{algorithm}
  \caption{Entropy Search}
  \label{alg}
  \begin{algorithmic}[1]
    \Procedure{EntropySearch}{$k,l=p(y\g f(x)),u,H,(\vx,\vy)$}
    \State $\tilde{\vx} \sim u(\vx,\vy)$ 
    \Comment{discretize using 
    measure $u$ (Section~\ref{sec:discr-repr-cont})}
    \State
    $[\vec\mu,\vec\Sigma,\Delta\vec{\mu}_x,\Delta\vec{\Sigma}_x]\gets$
    \Call{GP}{$k,l,\vx,\vy$} \Comment{infer function, innovation, from
      GP prior (\ref{sec:gauss-proc-meas})}
    \State $[\hat{q}_{\min}(\tilde{x}),\frac{\de
      \hat{q}_{\min}}{\de
      \mu},\frac{\de^2 \hat{q}_{\min}}{\de \mu\de \mu},
    \frac{\de \hat{q}_{\min}x}{\de \Sigma}]\gets$ 
    \Call{EP}{$\vec{\mu},\vec{\Sigma}$} 
    \Comment{approximate $\hat{p}_{\min}$ (\ref{sec:appr-pmin-with})}
    \If{H=0} \State\textbf{return} $q_{\min}$ 
    \Comment{At horizon, return belief for final decision} 
    \Else
    \State $x' \gets \argmin \langle \L\rangle_x$ \Comment{predict
      information gain; Eq. \eqref{eq:21}}
    \State $y' \gets$ \Call{Evaluate}{$f(x')$} \Comment{take measurement}
    \State \Call{EntropySearch}{$k,l,u,H-1,(\vx,\vy)\cup (x',y')$}
    \Comment{move to next evaluation}
    \EndIf
    \EndProcedure
  \end{algorithmic}
\end{algorithm}

Algorithm \ref{alg} shows pseudocode for Entropy Search. It takes as
input the prior, described by the kernel $k$, and the likelihood
$l=p(y\g f(x))$, as well as the discretization measure $u$ (which may
itself be a function of previous data, the Horizon $H$, and any
previously collected observations $(\vx,\vy)$. To choose where to
evaluate next, we first sample discretization points from $u$, then
calculate the current Gaussian belief over $f$ on the discretized
domain, along with its derivatives. We construct an approximation to
the belief over the minimum using Expectation Propagation, again with
derivatives. Finally, we construct a first order approximation on the
expected information gain from an evaluation at $x'$ and optimize
numerically. We evaluate $f$ at this location, then the cycle
repeats. Upon publication of this work, {\sc matlab} source code for
the algorithm and its sub-routines will be made available
online.

\section{Experiments}
\label{sec:loss-functions}

Figures in previous sections provided some intuition and anecdotal
evidence for the efficacy of the various approximations used by
Entropy Search. In this section, we compare the resulting algorithm to
two Gaussian process global optimization heuristics: Expected
Improvement, Probability of Improvement (Section
\ref{sec:relat-heur-gauss}), as well as to a continuous armed bandit
algorithm: GP-UCB \citep{srinivasgaussian}. For reference, we also
compare to a number of numerical optimization algorithms:
Trust-Region-Reflective
\citep{coleman1996interior,coleman1994convergence}, Active-Set
\citep{powell1978fast,powell1978convergence}, interior point
\citep{byrd1999interior,byrd2000trust,waltz2006interior}, and a
na\"ively projected version of the BFGS algorithm
\citep{broyden1965class,fletcher1970new,goldfarb1970family,shanno1970conditioning}.
We avoid implementation bias by using a uniform code framework for the
three Gaussian process-based algorithms, i.e.\ the algorithms share
code for the Gaussian process inference and only differ in the way
they calculate their utility. For the local numerical algorithms, we
used third party code: The projected BFGS method is based on code by
Carl Rasmussen\footnote{{\tt
    http://www.gaussianprocess.org/gpml/code/matlab/util/minimize.m},
  version using BFGS: personal communication}, the other methods come
from version 6.0 of the optimization toolbox of {\sc
  matlab}\footnote{\tt
  http://www.mathworks.de/help/toolbox/optim/rn/bsqj\_zi.html}.

In some communities, optimization algorithms are tested on
hand-crafted test functions. This runs the risk of introducing
bias. Instead, we compare our algorithms on a number of functions
sampled from a generative model. In the first experiment, the function
is sampled from the model used by the GP algorithms themselves. This
eliminates all model-mismatch issues and allows a direct comparison of
other GP optimizers to the probabilistic optimizer. In a second
experiment, the functions were sampled from a model strictly more
general than the model used by the algorithms, to show the effect of
model mismatch.

\subsection{Within-Model Comparison}
\label{sec:model-comparison}

The first experiment was carried out over the 2-dimensional unit
domain $I=[0,1]^2$. To generate test functions, 1000 function values
were jointly sampled from a Gaussian process with a squared-exponential
covariance function of length scale $\ell=0.1$ in each direction and
unit signal variance. The resulting posterior mean was used as the
test function. All algorithms had access to noisy evaluations of the
test functions. For the benefit of the numerical optimizers, noise was
kept relatively low: Gaussian with standard deviation
$\sigma=10^{-3}$. All algorithms were tested on the same set of 40
test functions, all Figures in this section are averages over those
sets of functions. It is nontrivial to provide error bars on these
average estimates, because the data sets have no parametric
distribution. But the regular structure of the plots, given that
individual experiments were drawn i.i.d., indicates that there is
little remaining stochastic error.

\begin{figure}
  \centering
   \beginpgfgraphicnamed{figures/fdist_wB_2D_SE-external}%
   \input{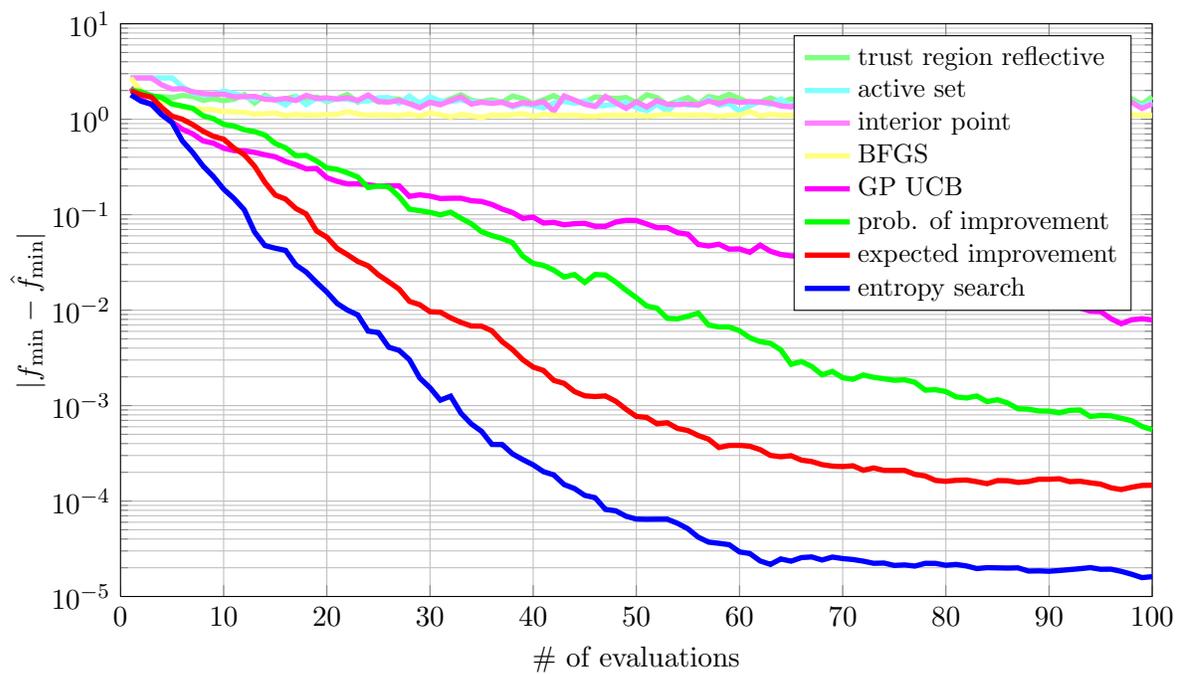}%
   \endpgfgraphicnamed%
 
  \caption{Distance of function value at optimizers' best guess for
    $\xmin$ from true global minimum. Log scale.}
  \label{fig:fval_onmodel}
\end{figure}

After each function evaluation, the algorithms were asked to return a
best guess for the minimum $\xmin$. For the local algorithms, this is
simply the point of their next evaluation. The Gaussian process based
methods returned the global minimum of the mean belief over the
function (found by local optimization with random restarts). Figure
\ref{fig:fval_onmodel} shows the difference between the global optimum
of the function and the function value at the reported best
guesses. Since the best guesses do not in general lie at a datapoint,
their quality can actually decrease during optimization. The most obvious
feature of this plot is that local optimization algorithms are not
adept at finding global minima, which is not surprising, but gives an
intuition for the difficulty of problems sampled from this generative
model. The plot shows a clear advantage for Entropy Search over its
competitors, even though the algorithm does not directly aim to
optimize this particular loss function. The flattening out of the
error of all three global optimizers toward the right is due to
evaluation noise (recall that evaluations include Gaussian noise of
standard deviation $10^{-3}$). Interestingly, Entropy Search flattens
out at an error almost an order of magnitude lower than that of the
nearest competitor, Expected Improvement. One possible explanation for
this behavior is a pathology in the classic heuristics: Both Expected
Improvement and Probability of Improvement require a ``current best
guess'' $\eta$, which has to be a point estimate, because proper
marginalization over an uncertain belief is not tractable. Due to
noise, it can thus happen that this best guess is overly optimistic,
and the algorithm then explores too aggressively in later stages.

\begin{figure}
  \centering
   \beginpgfgraphicnamed{figures/xdist_wB_2D_SE-external}%
   \input{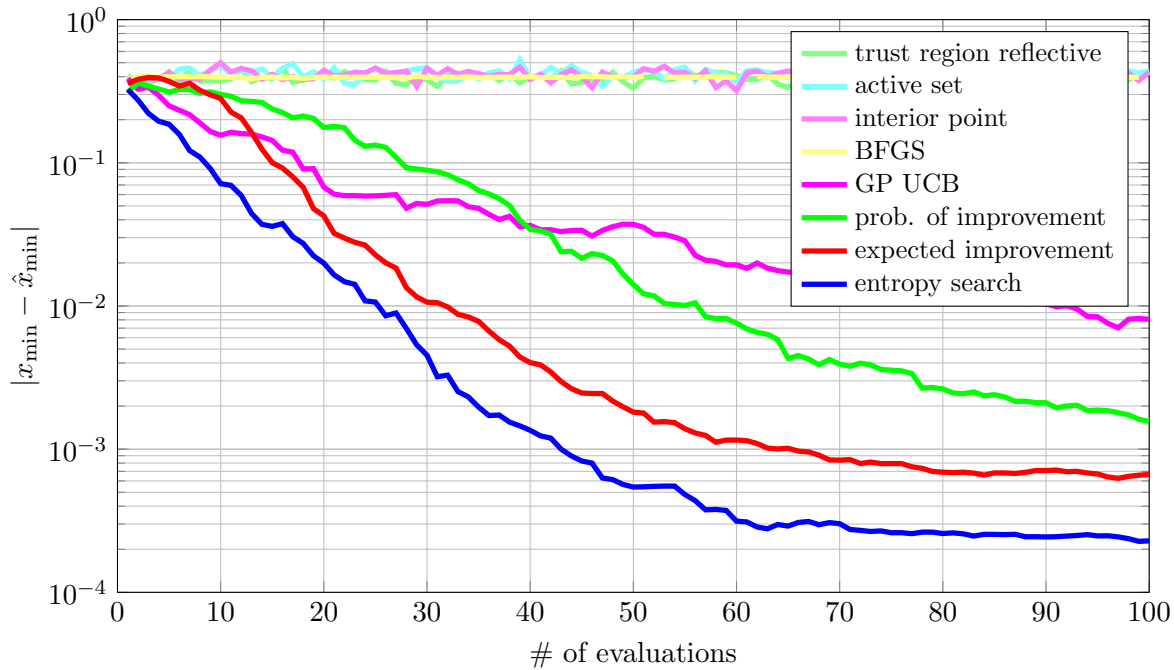}%
   \endpgfgraphicnamed%
 
  \caption{Euclidean distance of optimizers' best guess for $\xmin$
    from truth. Log scale.}
  \label{fig:xval_onmodel}
\end{figure}

Figure \ref{fig:xval_onmodel} shows data from the same experiments as
the previous figure, but plots Euclidean distance from the true
global optimum in input space, rather than in function value
space. The results from this view are qualitatively similar to those
shown in Figure \ref{fig:fval_onmodel}.

Since Entropy Search attempts to optimize information gain from
evaluations, one would also like to compare to algorithms on the
entropy loss function. However, this is challenging. First, the local
optimization algorithms provide no probabilistic model of the function
and can thus not provide this loss. But even for the optimization
algorithms based on Gaussian process measures, it is challenging to
evaluate this loss \emph{globally} with good resolution. The only
option is to approximately calculate entropy, using the very algorithm
introduced in this paper. Doing so amounts to a kind of circular
experiment that Entropy Search wins by definition, so we omit it here.

\begin{figure}
  \centering
   \beginpgfgraphicnamed{figures/regret_wB_2D_SE-external}%
%
%
\begin{tikzpicture}

\definecolor{mycolor1}{rgb}{1,0,1}

\begin{axis}[%
scale only axis,
width=0.9\textwidth,
height=0.5\textwidth,
xmin=0, xmax=100,
ymin=0, ymax=300,
xlabel={$$\# of evaluations$$},
ylabel={$$regret$$},
xmajorgrids,
ymajorgrids,
zmajorgrids,
legend entries={$$\small GP UCB$$,$$\small prob. of improvement$$,$$\small expected improvement$$,$$\small entropy search$$},
legend style={at={(0.03,0.97)},anchor=north west,nodes=right}]
\addplot [
color=mycolor1,
solid,
line width=2.0pt
]
coordinates{
 (1,2.71997)(2,5.1546)(3,7.54923)(4,9.85313)(5,12.1308)(6,14.6405)(7,16.9325)(8,19.4464)(9,22.0292)(10,24.6969)(11,27.3279)(12,29.7907)(13,32.3567)(14,34.9185)(15,37.6657)(16,40.1106)(17,42.6759)(18,45.043)(19,47.7598)(20,50.0457)(21,52.6712)(22,55.5826)(23,58.0907)(24,60.7079)(25,63.5411)(26,66.2838)(27,68.8509)(28,71.4044)(29,74.2328)(30,77.0314)(31,79.6307)(32,82.2051)(33,85.1339)(34,87.9585)(35,90.6709)(36,93.453)(37,96.2797)(38,99.0194)(39,101.648)(40,104.226)(41,106.913)(42,109.508)(43,112.278)(44,115.034)(45,117.365)(46,119.821)(47,122.644)(48,125.302)(49,127.885)(50,130.388)(51,133.016)(52,135.4)(53,138.151)(54,140.544)(55,142.895)(56,145.398)(57,147.931)(58,150.246)(59,152.915)(60,155.612)(61,158.066)(62,160.5)(63,162.819)(64,165.272)(65,167.917)(66,170.685)(67,173.334)(68,175.858)(69,178.55)(70,180.909)(71,183.365)(72,185.652)(73,188.116)(74,190.718)(75,193.198)(76,195.762)(77,197.871)(78,200.384)(79,202.697)(80,205.129)(81,207.86)(82,210.337)(83,212.903)(84,215.536)(85,218.102)(86,220.593)(87,223.149)(88,225.852)(89,228.371)(90,230.824)(91,233.362)(92,235.973)(93,238.1)(94,240.311)(95,242.727)(96,245.29)(97,247.435)(98,250.101)(99,252.51)(100,255.095) 
};

\addplot [
color=green,
solid,
line width=2.0pt
]
coordinates{
 (1,2.82138)(2,5.11606)(3,7.18473)(4,9.17819)(5,11.1583)(6,12.9663)(7,14.669)(8,16.3815)(9,17.9642)(10,19.4758)(11,20.8601)(12,22.2538)(13,23.5442)(14,24.8279)(15,26.0706)(16,27.3331)(17,28.6879)(18,29.9545)(19,31.1958)(20,32.3756)(21,33.5484)(22,34.9856)(23,36.2689)(24,37.6313)(25,38.9067)(26,40.1883)(27,41.7825)(28,43.0613)(29,44.1592)(30,45.2789)(31,46.6044)(32,48.059)(33,49.4519)(34,50.93)(35,52.6412)(36,54.3427)(37,55.8715)(38,57.5871)(39,59.0258)(40,60.4788)(41,61.913)(42,63.8162)(43,65.1638)(44,66.4242)(45,68.0079)(46,69.3217)(47,70.7716)(48,72.1772)(49,73.3598)(50,75.0367)(51,76.4729)(52,77.9807)(53,79.4546)(54,80.981)(55,82.4773)(56,83.8187)(57,85.4609)(58,86.8435)(59,88.5082)(60,89.8972)(61,91.5756)(62,93.2969)(63,94.9453)(64,96.504)(65,98.202)(66,99.6886)(67,100.987)(68,102.815)(69,104.527)(70,106.293)(71,107.909)(72,109.777)(73,111.581)(74,113.361)(75,115.392)(76,117.451)(77,119.214)(78,120.647)(79,122.497)(80,124.197)(81,125.888)(82,127.523)(83,129.42)(84,131.061)(85,133.042)(86,134.914)(87,136.904)(88,138.235)(89,140.07)(90,141.602)(91,143.215)(92,145.321)(93,147.157)(94,148.62)(95,150.426)(96,151.867)(97,153.501)(98,155.444)(99,157.172)(100,158.674) 
};

\addplot [
color=red,
solid,
line width=2.0pt
]
coordinates{
 (1,2.8393)(2,5.25908)(3,7.69606)(4,10.0175)(5,12.0533)(6,14.2846)(7,16.1279)(8,18.1676)(9,19.9697)(10,22.0236)(11,23.7053)(12,25.472)(13,27.3571)(14,29.1071)(15,30.7202)(16,32.373)(17,34.2495)(18,36.0679)(19,37.8938)(20,39.7062)(21,41.5049)(22,43.5476)(23,45.6847)(24,48.0132)(25,50.0338)(26,51.8476)(27,53.5667)(28,55.5834)(29,57.7566)(30,59.9629)(31,61.9623)(32,64.0782)(33,66.1487)(34,68.4234)(35,70.9485)(36,73.2285)(37,75.1516)(38,76.8503)(39,79.0091)(40,80.9841)(41,82.9631)(42,85.0941)(43,87.3491)(44,89.1517)(45,91.6612)(46,93.9295)(47,96.1958)(48,98.2994)(49,100.866)(50,103.272)(51,105.306)(52,107.52)(53,109.939)(54,112.268)(55,114.544)(56,116.883)(57,119.391)(58,121.89)(59,124.456)(60,126.762)(61,129.315)(62,131.925)(63,134.33)(64,136.756)(65,139.145)(66,141.209)(67,143.452)(68,145.94)(69,148.279)(70,150.586)(71,152.599)(72,154.791)(73,157.228)(74,159.657)(75,162.239)(76,164.671)(77,167.116)(78,169.329)(79,171.26)(80,173.494)(81,175.521)(82,177.794)(83,179.98)(84,182.245)(85,184.08)(86,186.413)(87,188.459)(88,190.769)(89,193.278)(90,195.743)(91,197.902)(92,200.222)(93,202.388)(94,204.78)(95,206.58)(96,208.981)(97,211.306)(98,213.596)(99,215.819)(100,217.919) 
};

\addplot [
color=blue,
solid,
line width=2.0pt
]
coordinates{
 (1,2.70707)(2,5.42609)(3,7.78211)(4,10.0182)(5,12.0261)(6,13.8997)(7,15.6102)(8,17.2038)(9,18.8919)(10,20.4947)(11,22.1752)(12,23.8409)(13,25.4054)(14,27.1106)(15,28.74)(16,30.4448)(17,32.1028)(18,33.876)(19,35.562)(20,37.4015)(21,39.0462)(22,40.8828)(23,42.5093)(24,44.1938)(25,45.8708)(26,47.2331)(27,48.8289)(28,50.6438)(29,52.4052)(30,54.072)(31,56.0171)(32,57.8417)(33,59.5031)(34,61.0064)(35,62.6729)(36,64.3858)(37,65.9725)(38,67.5958)(39,69.6429)(40,71.5423)(41,73.2258)(42,75.3169)(43,77.0873)(44,78.9885)(45,80.9328)(46,82.8363)(47,84.5977)(48,86.7132)(49,88.3679)(50,90.0706)(51,92.0368)(52,93.9749)(53,96.0866)(54,98.3405)(55,100.444)(56,102.61)(57,104.508)(58,106.662)(59,108.403)(60,110.559)(61,112.421)(62,114.34)(63,116.539)(64,118.642)(65,120.779)(66,122.979)(67,125.01)(68,127.105)(69,129.144)(70,131.171)(71,133.499)(72,135.556)(73,137.943)(74,140.139)(75,142.287)(76,144.213)(77,146.531)(78,148.521)(79,150.786)(80,152.79)(81,155.02)(82,156.917)(83,158.973)(84,161.098)(85,163.276)(86,165.475)(87,167.531)(88,169.609)(89,172.036)(90,174.034)(91,176.177)(92,178.449)(93,180.924)(94,183)(95,185.273)(96,187.464)(97,189.594)(98,191.882)(99,194.172)(100,196.223) 
};

\end{axis}
\end{tikzpicture}%
   \endpgfgraphicnamed%
 
  \caption{Regret as a function of number of evaluations.}
  \label{fig:regret}
\end{figure}
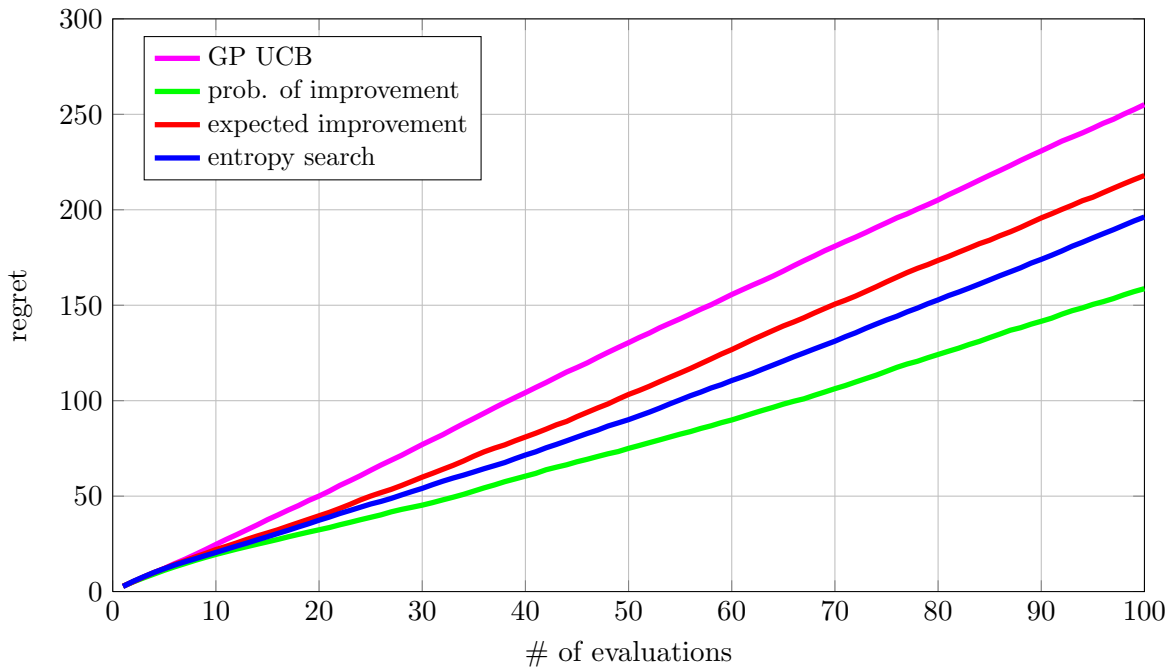

We pointed out in Section \ref{sec:diff-band-sett} that the bandit
setting differs considerably from the kind of optimization discussed
in this paper, because bandit algorithms try to minimize regret,
rather than improve an estimate of the function's optimum. To clarify
this point further, Figure \ref{fig:regret} shows the regret
\begin{equation}
  \label{eq:29}
  r(T) = \sum_{t=1} ^T [y_t - \fmin]
\end{equation}
for each of the algorithms. Notice that probability of improvement,
which performs worst among the global algorithms as seen from the
previous two measures of performance, achieves the lowest regret. The
intuition here is that this heuristic focusses evaluations on regions
known to give low function values. In contrast, the actual value of
the function \emph{at the evaluation point} has no special role in
Entropy Search. The utility of an evaluation point only depends on its
expected effect on knowledge about the minimum of the function.

Surprisingly, the one algorithm explicitly designed to achieve low
regret, GP-UCB, performs worst in this comparison. This algorithm
chooses evaluation points according to \citep{srinivasgaussian}
\begin{equation}
  \label{eq:30}
  x_{\text{next}} = \argmin_x [ \mu(x) - \beta^{1/2} \sigma(x) ]\qquad
\text{where} \qquad \beta = 4(D+1)\log T + C(k,\delta)
\end{equation}
with $T$, the number of previous evaluations, $D$, the dimensionality
of the input domain, and $C(k,\delta)$ is a constant that depends on
some analytic properties of the kernel $k$ and a free parameter,
$0<\delta<1$. We found it hard to find a good setting for this
$\delta$, which clearly has influence on the algorithm's
performance. The results shown here represent the best performance
over a set of 4 experiments with different choices for $\delta$. They
appear to be slightly worse than, but comparable to the empirical
performance reported by the original paper on this algorithm
\citep[][Figure 5a]{srinivasgaussian}.

\subsection{Out-of-Model Comparison}
\label{sec:out-model-comparison}

\begin{figure}
  \centering
   \beginpgfgraphicnamed{figures/SE_example-external}%
%
%
\begin{tikzpicture}

\begin{axis}[%
name=plot1,
scale only axis,
width=0.4\textwidth,
height=0.4\textwidth,
y dir=reverse,
xmin=-0.0025, xmax=1.0025,
ymin=-0.0025, ymax=1.0025,
axis on top]
\addplot graphics [xmin=-2.500000e-03, xmax=1.002500e+00, ymin=-2.500000e-03, ymax=1.002500e+00] {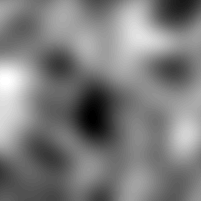};
\end{axis}

\begin{axis}[%
axis on top,
at=(plot1.right of south east), anchor=left of south west,
width=0.027027\textwidth, height=0.4\textwidth,
scale only axis,
xmin=0, xmax=1,
ymin=-2.95623, ymax=3.35166,
xtick=\empty, yticklabel pos=right,
xticklabels={\empty}]
\addplot graphics [xmin=0, xmax=1, ymin=-2.956227e+00, ymax=3.351658e+00] {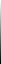};
\end{axis}
\end{tikzpicture}%
   \endpgfgraphicnamed%
   \beginpgfgraphicnamed{figures/RQ_example-external}%
%
%
\begin{tikzpicture}

\begin{axis}[%
name=plot1,
scale only axis,
width=0.4\textwidth,
height=0.4\textwidth,
y dir=reverse,
xmin=-0.0025, xmax=1.0025,
ymin=-0.0025, ymax=1.0025,
axis on top]
\addplot graphics [xmin=-2.500000e-03, xmax=1.002500e+00, ymin=-2.500000e-03, ymax=1.002500e+00] {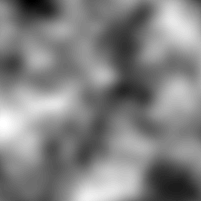};
\end{axis}

\begin{axis}[%
axis on top,
at=(plot1.right of south east), anchor=left of south west,
width=0.027027\textwidth, height=0.4\textwidth,
scale only axis,
xmin=0, xmax=1,
ymin=-1.91618, ymax=3.25746,
xtick=\empty, yticklabel pos=right,
xticklabels={\empty}]
\addplot graphics [xmin=0, xmax=1, ymin=-1.916183e+00, ymax=3.257457e+00] {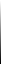};
\end{axis}
\end{tikzpicture}%
   \endpgfgraphicnamed%
 
  \caption{{\bf Left:} A sample from the GP prior with squared
    exponential kernel used in the on-model experiments of Section
    \ref{sec:model-comparison}. {\bf Right:} Sample from prior with
    the rational quadratic kernel used for the out of model comparison
    of Section \ref{sec:out-model-comparison}.}
  \label{fig:SE_RQ}
\end{figure}

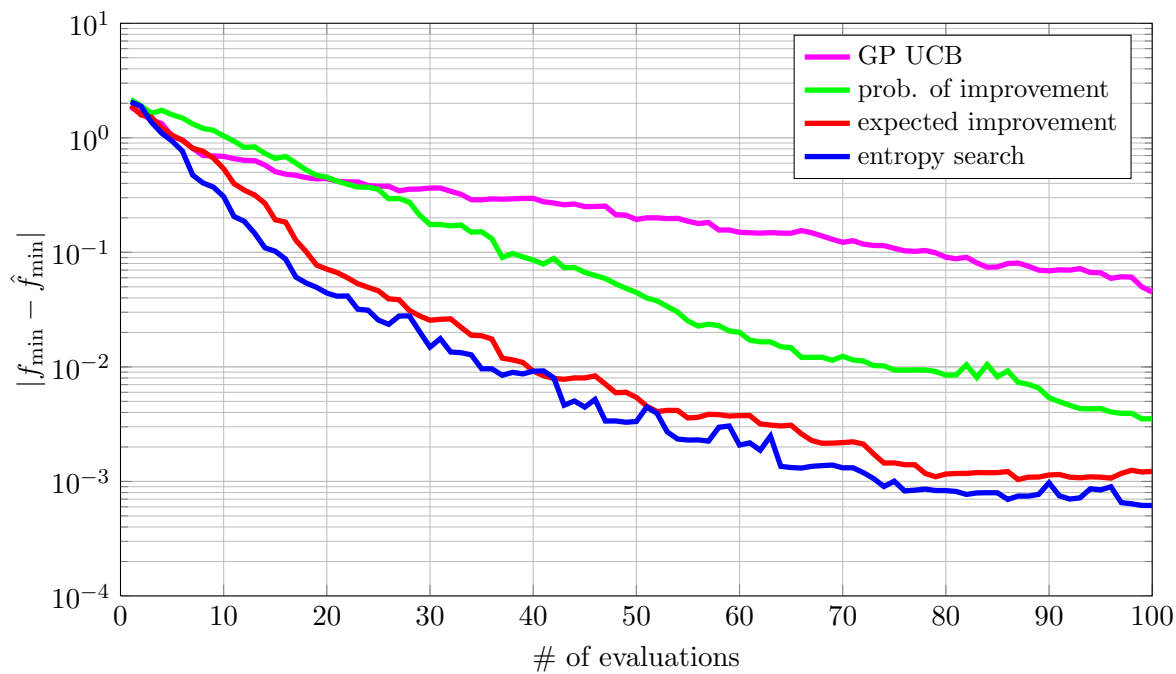
\begin{figure}
  \centering
   \beginpgfgraphicnamed{figures/fdist_wB_2D_RQ-external}%
%
%
\begin{tikzpicture}

\definecolor{mycolor1}{rgb}{1,0,1}

\begin{semilogyaxis}[%
scale only axis,
width=0.9\textwidth,
height=0.5\textwidth,
xmin=0, xmax=100,
ymin=0.0001, ymax=10,
xlabel={$$\# of evaluations$$},
ylabel={$|f _{\min} - \hat{f}_{\min}|$},
xmajorgrids,
ymajorgrids,
yminorgrids,
zmajorgrids,
legend entries={$$\small GP UCB$$,$$\small prob. of improvement$$,$$\small expected improvement$$,$$\small entropy search$$},
legend style={nodes=right}]
\addplot [
color=mycolor1,
solid,
line width=2.0pt
]
coordinates{
 (1,1.85173)(2,1.63228)(3,1.42211)(4,1.33226)(5,1.02821)(6,0.956878)(7,0.809698)(8,0.700502)(9,0.693757)(10,0.68799)(11,0.65726)(12,0.635898)(13,0.631699)(14,0.578964)(15,0.50576)(16,0.481355)(17,0.470092)(18,0.449275)(19,0.437565)(20,0.441233)(21,0.417007)(22,0.412199)(23,0.411743)(24,0.383267)(25,0.377605)(26,0.377828)(27,0.344313)(28,0.355753)(29,0.356834)(30,0.36424)(31,0.364214)(32,0.341863)(33,0.321299)(34,0.288015)(35,0.287258)(36,0.292956)(37,0.290914)(38,0.29319)(39,0.295408)(40,0.296207)(41,0.276514)(42,0.269439)(43,0.260063)(44,0.264302)(45,0.250695)(46,0.251321)(47,0.253321)(48,0.213217)(49,0.210994)(50,0.193645)(51,0.200236)(52,0.200182)(53,0.196969)(54,0.198112)(55,0.186852)(56,0.178541)(57,0.182304)(58,0.156504)(59,0.156721)(60,0.149705)(61,0.148192)(62,0.147179)(63,0.148413)(64,0.147329)(65,0.146902)(66,0.154989)(67,0.148234)(68,0.138988)(69,0.129833)(70,0.122356)(71,0.126391)(72,0.117933)(73,0.114717)(74,0.114535)(75,0.108492)(76,0.103151)(77,0.102013)(78,0.103761)(79,0.0991709)(80,0.0906742)(81,0.0879565)(82,0.0905889)(83,0.0809467)(84,0.0741409)(85,0.0745824)(86,0.0798543)(87,0.0806402)(88,0.0756342)(89,0.0699083)(90,0.0688904)(91,0.0703412)(92,0.0699872)(93,0.0720587)(94,0.0667633)(95,0.0663372)(96,0.0592778)(97,0.0611699)(98,0.0607594)(99,0.0499789)(100,0.0450254) 
};

\addplot [
color=green,
solid,
line width=2.0pt
]
coordinates{
 (1,2.15421)(2,1.90462)(3,1.63274)(4,1.73942)(5,1.58998)(6,1.48905)(7,1.32185)(8,1.20812)(9,1.16315)(10,1.03892)(11,0.936504)(12,0.822993)(13,0.834216)(14,0.73145)(15,0.66095)(16,0.687252)(17,0.602378)(18,0.524112)(19,0.470511)(20,0.452943)(21,0.417574)(22,0.393569)(23,0.371363)(24,0.369891)(25,0.354961)(26,0.293565)(27,0.295424)(28,0.2742)(29,0.211011)(30,0.175096)(31,0.174832)(32,0.170147)(33,0.173484)(34,0.150038)(35,0.151419)(36,0.131471)(37,0.0902327)(38,0.0978531)(39,0.0913453)(40,0.0856431)(41,0.0790286)(42,0.0883927)(43,0.0736052)(44,0.0738157)(45,0.0669351)(46,0.0627889)(47,0.059076)(48,0.0530985)(49,0.0484984)(50,0.0447083)(51,0.0399597)(52,0.0377743)(53,0.0336782)(54,0.030124)(55,0.0251247)(56,0.0227135)(57,0.0236131)(58,0.0228941)(59,0.020577)(60,0.0200449)(61,0.0170921)(62,0.0165578)(63,0.0165415)(64,0.0150068)(65,0.0146689)(66,0.0120965)(67,0.0121134)(68,0.0121316)(69,0.0114151)(70,0.0123793)(71,0.0114973)(72,0.0112512)(73,0.0102592)(74,0.0102146)(75,0.0093969)(76,0.00936984)(77,0.00940035)(78,0.00939615)(79,0.00912905)(80,0.00848297)(81,0.00849593)(82,0.0104314)(83,0.00808323)(84,0.0104885)(85,0.00820341)(86,0.00922672)(87,0.00733573)(88,0.00704573)(89,0.00653055)(90,0.00538677)(91,0.0049736)(92,0.00461759)(93,0.00432397)(94,0.00429789)(95,0.00432064)(96,0.00404346)(97,0.00393138)(98,0.00393493)(99,0.00351687)(100,0.00352761) 
};

\addplot [
color=red,
solid,
line width=2.0pt
]
coordinates{
 (1,1.91965)(2,1.58244)(3,1.51137)(4,1.21676)(5,1.05563)(6,0.950804)(7,0.808624)(8,0.76408)(9,0.665012)(10,0.537411)(11,0.397783)(12,0.348944)(13,0.316703)(14,0.268113)(15,0.192338)(16,0.183637)(17,0.127415)(18,0.101236)(19,0.0775194)(20,0.0712976)(21,0.0666062)(22,0.059957)(23,0.0532044)(24,0.049508)(25,0.0460875)(26,0.0392351)(27,0.0386514)(28,0.031055)(29,0.0278524)(30,0.0255382)(31,0.0260048)(32,0.0262807)(33,0.022187)(34,0.018913)(35,0.0187045)(36,0.0175067)(37,0.0119194)(38,0.0115173)(39,0.0109376)(40,0.00919019)(41,0.00837284)(42,0.00792027)(43,0.00779893)(44,0.00801025)(45,0.00801136)(46,0.00834313)(47,0.00707028)(48,0.00594447)(49,0.00600693)(50,0.00541546)(51,0.00454817)(52,0.00407137)(53,0.00417113)(54,0.00416617)(55,0.00358071)(56,0.00363261)(57,0.00385492)(58,0.0038297)(59,0.00371979)(60,0.00375803)(61,0.00375658)(62,0.00317912)(63,0.003108)(64,0.00304934)(65,0.00309736)(66,0.00260142)(67,0.00227648)(68,0.00215483)(69,0.00215943)(70,0.0021825)(71,0.00221736)(72,0.00212147)(73,0.00173543)(74,0.00144747)(75,0.00145159)(76,0.00140063)(77,0.00139869)(78,0.00117499)(79,0.00109829)(80,0.00116047)(81,0.00117376)(82,0.00117486)(83,0.00119516)(84,0.00119157)(85,0.00119378)(86,0.00122115)(87,0.0010391)(88,0.00108972)(89,0.00109234)(90,0.00113826)(91,0.00114937)(92,0.00109021)(93,0.00107726)(94,0.00109682)(95,0.00108991)(96,0.001069)(97,0.00117443)(98,0.00125131)(99,0.00120884)(100,0.00122225) 
};

\addplot [
color=blue,
solid,
line width=2.0pt
]
coordinates{
 (1,2.05987)(2,1.88215)(3,1.3669)(4,1.09593)(5,0.936155)(6,0.765479)(7,0.472408)(8,0.403326)(9,0.371791)(10,0.306719)(11,0.205709)(12,0.18643)(13,0.147057)(14,0.109503)(15,0.102244)(16,0.0873175)(17,0.0607906)(18,0.0539737)(19,0.0495109)(20,0.0440866)(21,0.0414239)(22,0.0415987)(23,0.0318711)(24,0.0312402)(25,0.0256039)(26,0.0235165)(27,0.0278026)(28,0.0278837)(29,0.0202377)(30,0.0148702)(31,0.0176415)(32,0.0134799)(33,0.0133066)(34,0.0127438)(35,0.00964034)(36,0.00962041)(37,0.00845102)(38,0.00895581)(39,0.00868504)(40,0.00917576)(41,0.00922557)(42,0.00803001)(43,0.00462598)(44,0.00503692)(45,0.00445496)(46,0.00519842)(47,0.00336911)(48,0.00337155)(49,0.00328501)(50,0.00333608)(51,0.00445021)(52,0.00393826)(53,0.00270237)(54,0.00234535)(55,0.00229803)(56,0.0023046)(57,0.00224919)(58,0.00297177)(59,0.00304817)(60,0.00207742)(61,0.00217205)(62,0.00187218)(63,0.00248581)(64,0.0013536)(65,0.00132258)(66,0.00130852)(67,0.00135446)(68,0.00137364)(69,0.0013903)(70,0.00131503)(71,0.00131581)(72,0.00119376)(73,0.0010586)(74,0.000902805)(75,0.00100545)(76,0.000825723)(77,0.000838907)(78,0.000858063)(79,0.000832303)(80,0.000831676)(81,0.000817512)(82,0.000769818)(83,0.000794006)(84,0.000796628)(85,0.000796198)(86,0.00069652)(87,0.000745661)(88,0.000744597)(89,0.000771466)(90,0.00097325)(91,0.000748972)(92,0.000701999)(93,0.000718347)(94,0.00086237)(95,0.000843791)(96,0.000898467)(97,0.000653215)(98,0.000638263)(99,0.000617087)(100,0.000616421) 
};

\end{semilogyaxis}
\end{tikzpicture}%
   \endpgfgraphicnamed%
 
  \caption{Function value error, off-model tasks.}
  \label{fig:fval_offmodel}
\end{figure}

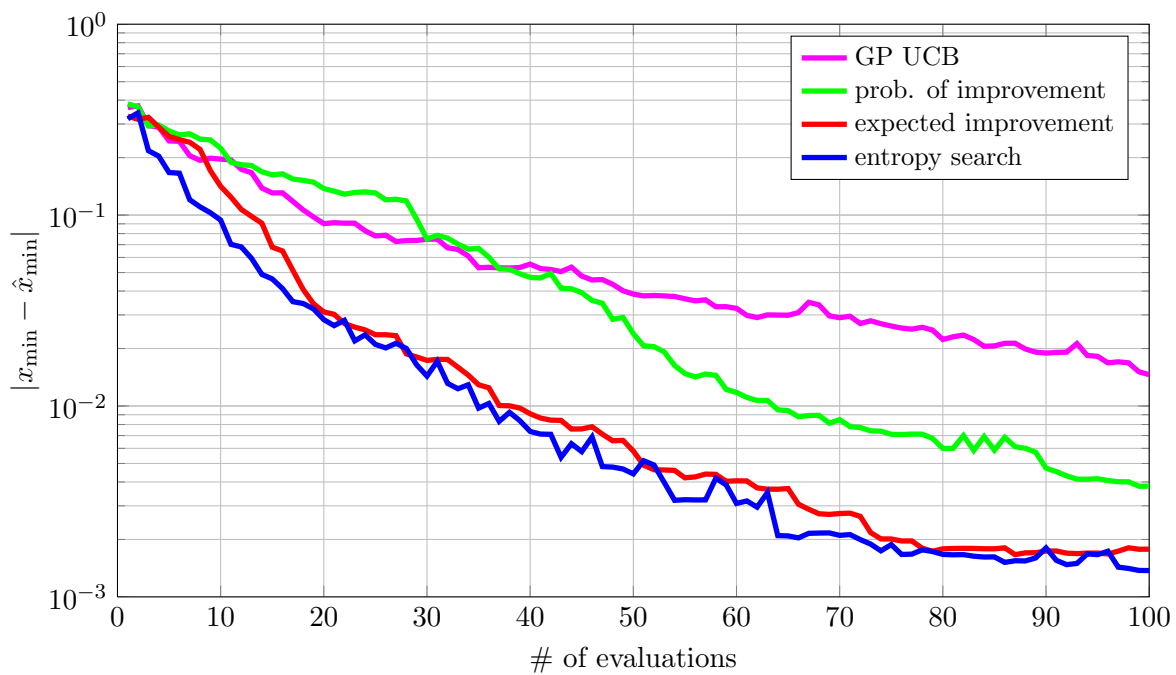
\begin{figure}
  \centering
   \beginpgfgraphicnamed{figures/xdist_wB_2D_RQ-external}%
%
%
\begin{tikzpicture}

\definecolor{mycolor1}{rgb}{1,0,1}

\begin{semilogyaxis}[%
scale only axis,
width=0.9\textwidth,
height=0.5\textwidth,
xmin=0, xmax=100,
ymin=0.001, ymax=1,
xlabel={$$\# of evaluations$$},
ylabel={$|x_{\min} - \hat{x}_{\min}|$},
xmajorgrids,
ymajorgrids,
yminorgrids,
zmajorgrids,
legend entries={$$\small GP UCB$$,$$\small prob. of improvement$$,$$\small expected improvement$$,$$\small entropy search$$},
legend style={nodes=right}]
\addplot [
color=mycolor1,
solid,
line width=2.0pt
]
coordinates{
 (1,0.366061)(2,0.373139)(3,0.294443)(4,0.28899)(5,0.244638)(6,0.243543)(7,0.2047)(8,0.193596)(9,0.1987)(10,0.196518)(11,0.194448)(12,0.173725)(13,0.166694)(14,0.138496)(15,0.130695)(16,0.130752)(17,0.117993)(18,0.106471)(19,0.0977879)(20,0.090189)(21,0.0912218)(22,0.0906951)(23,0.090689)(24,0.0827641)(25,0.0777898)(26,0.0784636)(27,0.0727164)(28,0.0734804)(29,0.0736396)(30,0.0747681)(31,0.0748024)(32,0.0673986)(33,0.0660743)(34,0.0610543)(35,0.0529986)(36,0.0532709)(37,0.052909)(38,0.0528317)(39,0.0531503)(40,0.0553715)(41,0.0524759)(42,0.0518621)(43,0.0505928)(44,0.0533654)(45,0.047931)(46,0.0457389)(47,0.045961)(48,0.0434145)(49,0.0401521)(50,0.0385805)(51,0.037746)(52,0.0379241)(53,0.0377241)(54,0.0374872)(55,0.0364489)(56,0.035556)(57,0.0359736)(58,0.0331027)(59,0.0331354)(60,0.0324763)(61,0.0298515)(62,0.0290617)(63,0.0300044)(64,0.0299453)(65,0.029886)(66,0.0308518)(67,0.034926)(68,0.0338674)(69,0.0296557)(70,0.0290097)(71,0.0295234)(72,0.0270435)(73,0.0279502)(74,0.0269892)(75,0.0262299)(76,0.0255932)(77,0.0252605)(78,0.0258403)(79,0.0249908)(80,0.0223013)(81,0.0230505)(82,0.0235181)(83,0.0223039)(84,0.020542)(85,0.0206465)(86,0.0212986)(87,0.0213171)(88,0.0198907)(89,0.0191577)(90,0.0189282)(91,0.0190841)(92,0.0191508)(93,0.0212361)(94,0.0183967)(95,0.01821)(96,0.0168277)(97,0.017041)(98,0.0168316)(99,0.0151098)(100,0.0145912) 
};

\addplot [
color=green,
solid,
line width=2.0pt
]
coordinates{
 (1,0.382576)(2,0.368578)(3,0.292999)(4,0.296332)(5,0.27604)(6,0.262782)(7,0.266629)(8,0.250157)(9,0.24771)(10,0.223951)(11,0.189043)(12,0.183619)(13,0.181842)(14,0.168585)(15,0.162428)(16,0.164472)(17,0.154516)(18,0.151985)(19,0.148862)(20,0.137747)(21,0.133517)(22,0.128643)(23,0.131395)(24,0.132291)(25,0.130739)(26,0.120204)(27,0.121026)(28,0.118648)(29,0.0951992)(30,0.0749835)(31,0.0783793)(32,0.0757206)(33,0.0705337)(34,0.0663665)(35,0.0669606)(36,0.0601629)(37,0.0522323)(38,0.0519805)(39,0.0492985)(40,0.0472031)(41,0.046831)(42,0.0494304)(43,0.0413905)(44,0.0411156)(45,0.039264)(46,0.035713)(47,0.0345625)(48,0.0284469)(49,0.029135)(50,0.0239299)(51,0.0206584)(52,0.0204721)(53,0.0192001)(54,0.0162993)(55,0.0148038)(56,0.0142041)(57,0.0147055)(58,0.0144698)(59,0.0122202)(60,0.0117842)(61,0.0110982)(62,0.010686)(63,0.0106954)(64,0.00954747)(65,0.0094534)(66,0.0087917)(67,0.00891415)(68,0.00893379)(69,0.00812736)(70,0.00848635)(71,0.00780028)(72,0.00771384)(73,0.00741594)(74,0.00738267)(75,0.00710539)(76,0.00708981)(77,0.00712667)(78,0.00712264)(79,0.00676769)(80,0.00599584)(81,0.00600082)(82,0.00700247)(83,0.0058537)(84,0.00691875)(85,0.00587083)(86,0.00685403)(87,0.00612153)(88,0.00600573)(89,0.0057292)(90,0.00472318)(91,0.00453523)(92,0.00430691)(93,0.00413965)(94,0.00413329)(95,0.0041588)(96,0.00407044)(97,0.00400959)(98,0.00400875)(99,0.00380192)(100,0.00380254) 
};

\addplot [
color=red,
solid,
line width=2.0pt
]
coordinates{
 (1,0.329815)(2,0.317193)(3,0.32517)(4,0.289004)(5,0.258355)(6,0.248387)(7,0.240778)(8,0.221017)(9,0.171119)(10,0.141471)(11,0.124128)(12,0.107136)(13,0.0984458)(14,0.0906118)(15,0.0679154)(16,0.0649068)(17,0.0513329)(18,0.0407873)(19,0.0343228)(20,0.0311043)(21,0.0302179)(22,0.0270355)(23,0.0258856)(24,0.0250451)(25,0.0236222)(26,0.0236162)(27,0.0233618)(28,0.0187441)(29,0.0180668)(30,0.0173394)(31,0.0175468)(32,0.0175263)(33,0.0159588)(34,0.0145183)(35,0.0129123)(36,0.0124657)(37,0.0100586)(38,0.010021)(39,0.00975855)(40,0.00910755)(41,0.00865116)(42,0.00843058)(43,0.00840734)(44,0.00757599)(45,0.00758485)(46,0.00778792)(47,0.00712045)(48,0.00657342)(49,0.00660408)(50,0.00580985)(51,0.00489799)(52,0.00463617)(53,0.00462305)(54,0.0046004)(55,0.00420619)(56,0.00425162)(57,0.00439944)(58,0.00438547)(59,0.00403281)(60,0.00405598)(61,0.00404932)(62,0.00372476)(63,0.00366792)(64,0.00366458)(65,0.00369571)(66,0.00305899)(67,0.00287829)(68,0.00272557)(69,0.00270572)(70,0.00273392)(71,0.00274668)(72,0.00264183)(73,0.00217369)(74,0.00201272)(75,0.00201136)(76,0.00196383)(77,0.00196417)(78,0.00179944)(79,0.00173464)(80,0.00178636)(81,0.00179361)(82,0.00179674)(83,0.00179532)(84,0.00178734)(85,0.00178698)(86,0.00180739)(87,0.00166493)(88,0.00170179)(89,0.00170449)(90,0.00173892)(91,0.00174031)(92,0.00169494)(93,0.00168581)(94,0.00169658)(95,0.00169693)(96,0.00167909)(97,0.00173662)(98,0.00180843)(99,0.00177724)(100,0.00177901) 
};

\addplot [
color=blue,
solid,
line width=2.0pt
]
coordinates{
 (1,0.319732)(2,0.343298)(3,0.217511)(4,0.203565)(5,0.167051)(6,0.165761)(7,0.120391)(8,0.110466)(9,0.102937)(10,0.0941365)(11,0.0702685)(12,0.0681257)(13,0.0593088)(14,0.048877)(15,0.0463198)(16,0.0412127)(17,0.0351902)(18,0.0344321)(19,0.0322584)(20,0.0283455)(21,0.0263577)(22,0.0281035)(23,0.0219364)(24,0.0235918)(25,0.0210218)(26,0.0201516)(27,0.0213066)(28,0.0200175)(29,0.0164257)(30,0.0143464)(31,0.0171384)(32,0.0131333)(33,0.0123135)(34,0.0129152)(35,0.00974885)(36,0.0103109)(37,0.00831313)(38,0.00926337)(39,0.00837967)(40,0.00735925)(41,0.00712832)(42,0.00710755)(43,0.00538522)(44,0.00634675)(45,0.00576679)(46,0.00687694)(47,0.00481572)(48,0.00478626)(49,0.00467747)(50,0.00440905)(51,0.00516744)(52,0.00490975)(53,0.00397517)(54,0.00320612)(55,0.00322938)(56,0.00322091)(57,0.00322226)(58,0.00418363)(59,0.00386353)(60,0.00308707)(61,0.003182)(62,0.00295277)(63,0.00352196)(64,0.00209442)(65,0.00208736)(66,0.00203791)(67,0.00215305)(68,0.00216056)(69,0.0021637)(70,0.0021008)(71,0.00212133)(72,0.00199662)(73,0.00189166)(74,0.0017433)(75,0.00188004)(76,0.00166863)(77,0.00167597)(78,0.00176485)(79,0.00172936)(80,0.00166947)(81,0.00166182)(82,0.00166574)(83,0.00163372)(84,0.00161827)(85,0.00162051)(86,0.00151645)(87,0.0015468)(88,0.00154387)(89,0.00160059)(90,0.00180653)(91,0.00155432)(92,0.00147604)(93,0.00149768)(94,0.00168094)(95,0.00166435)(96,0.00173511)(97,0.00143179)(98,0.00140989)(99,0.0013765)(100,0.0013738) 
};

\end{semilogyaxis}
\end{tikzpicture}%
   \endpgfgraphicnamed%
 
  \caption{Error on $\xmin$, off-model tasks.}
  \label{fig:xval_offmodel}
\end{figure}

\begin{figure}
  \centering
   \beginpgfgraphicnamed{figures/regret_2D_wB_RQ-external}%
%
%
\begin{tikzpicture}

\definecolor{mycolor1}{rgb}{1,0,1}

\begin{axis}[%
scale only axis,
width=0.9\textwidth,
height=0.5\textwidth,
xmin=0, xmax=100,
ymin=0, ymax=220,
xlabel={$$\# of evaluations$$},
ylabel={$$regret$$},
xmajorgrids,
ymajorgrids,
zmajorgrids,
legend entries={$$\small GP UCB$$,$$\small prob. of improvement$$,$$\small expected improvement$$,$$\small entropy search$$},
legend style={at={(0.03,0.97)},anchor=north west,nodes=right}]
\addplot [
color=mycolor1,
solid,
line width=2.0pt
]
coordinates{
 (1,2.47467)(2,5.04007)(3,7.49638)(4,9.71201)(5,12.0087)(6,14.352)(7,16.4483)(8,18.6183)(9,20.8349)(10,23.0815)(11,25.3773)(12,27.7756)(13,29.9925)(14,32.7136)(15,35.154)(16,37.6964)(17,39.9933)(18,42.3762)(19,44.7414)(20,47.3242)(21,50.0766)(22,53.0094)(23,55.647)(24,57.9999)(25,60.5785)(26,63.3887)(27,66.0856)(28,68.9007)(29,71.7041)(30,74.525)(31,77.4262)(32,80.1617)(33,82.8583)(34,85.3171)(35,87.7982)(36,90.4437)(37,92.9785)(38,95.627)(39,98.0697)(40,100.633)(41,103.293)(42,105.947)(43,108.4)(44,110.837)(45,113.245)(46,115.541)(47,118.061)(48,120.464)(49,122.762)(50,125.445)(51,128.066)(52,130.652)(53,133.473)(54,136.109)(55,138.46)(56,140.769)(57,143.29)(58,145.713)(59,148.087)(60,150.852)(61,153.319)(62,156.084)(63,158.469)(64,161.204)(65,163.736)(66,166.582)(67,168.736)(68,171.576)(69,173.934)(70,176.55)(71,179.135)(72,181.539)(73,184.095)(74,186.938)(75,189.114)(76,191.628)(77,194.16)(78,196.544)(79,198.829)(80,201.289)(81,203.626)(82,205.865)(83,208.46)(84,210.619)(85,213.132)(86,215.563)(87,217.845)(88,219.955)(88.762,221.7) 
};

\addplot [
color=green,
solid,
line width=2.0pt
]
coordinates{
 (1,2.47459)(2,4.7435)(3,6.97081)(4,8.98461)(5,10.8424)(6,12.639)(7,14.46)(8,16.0827)(9,17.5749)(10,19.0236)(11,20.4009)(12,21.8406)(13,23.2877)(14,24.5709)(15,25.7828)(16,27.0298)(17,28.2395)(18,29.6085)(19,31.2865)(20,32.8775)(21,34.4773)(22,36.1444)(23,37.8813)(24,39.383)(25,40.6657)(26,42.0894)(27,43.6684)(28,45.326)(29,47.0073)(30,48.3135)(31,49.8129)(32,51.4571)(33,52.7667)(34,54.5565)(35,55.8342)(36,57.2654)(37,58.4269)(38,59.6377)(39,61.1556)(40,62.6492)(41,64.5033)(42,65.8864)(43,67.7396)(44,69.4441)(45,71.0545)(46,72.842)(47,74.4262)(48,75.9955)(49,77.7463)(50,79.1702)(51,80.8778)(52,82.2108)(53,84.1017)(54,85.7834)(55,87.5779)(56,89.027)(57,91.0904)(58,93.156)(59,95.0204)(60,96.6023)(61,98.4081)(62,100.228)(63,102.216)(64,103.992)(65,106.004)(66,107.806)(67,109.407)(68,111.105)(69,112.445)(70,114.434)(71,116.066)(72,117.619)(73,119.716)(74,121.187)(75,122.949)(76,124.742)(77,126.479)(78,128.519)(79,130.095)(80,131.516)(81,133.518)(82,135.268)(83,137.206)(84,138.986)(85,140.611)(86,142.229)(87,144.014)(88,146.056)(89,147.77)(90,149.367)(91,151.123)(92,152.685)(93,154.443)(94,156.457)(95,158.363)(96,159.896)(97,162.014)(98,163.847)(99,165.559)(100,167.523) 
};

\addplot [
color=red,
solid,
line width=2.0pt
]
coordinates{
 (1,2.69181)(2,5.13041)(3,7.27816)(4,9.41229)(5,11.4506)(6,13.4089)(7,15.882)(8,18.0175)(9,20.0631)(10,21.9151)(11,23.8016)(12,25.6736)(13,27.749)(14,29.5625)(15,31.4903)(16,33.1987)(17,34.7409)(18,36.3093)(19,38.0773)(20,39.8807)(21,41.7323)(22,43.7001)(23,45.2187)(24,47.3755)(25,49.4416)(26,51.3164)(27,53.4339)(28,55.4779)(29,57.4141)(30,59.5703)(31,61.65)(32,63.8982)(33,66.0897)(34,68.0493)(35,70.6886)(36,73.1008)(37,75.0904)(38,77.1807)(39,79.5069)(40,81.7644)(41,83.9601)(42,86.1843)(43,88.2654)(44,90.3907)(45,93.0452)(46,95.5231)(47,97.7996)(48,99.986)(49,102.445)(50,104.672)(51,107.035)(52,109.546)(53,111.494)(54,113.795)(55,116.026)(56,118.399)(57,120.751)(58,123.28)(59,125.775)(60,128.039)(61,130.3)(62,132.702)(63,135.128)(64,137.484)(65,139.65)(66,141.772)(67,143.639)(68,146.212)(69,148.416)(70,150.354)(71,152.487)(72,154.877)(73,157.159)(74,159.134)(75,161.183)(76,163.278)(77,165.589)(78,167.87)(79,170.087)(80,172.078)(81,174.289)(82,176.757)(83,179.253)(84,181.666)(85,183.87)(86,186.174)(87,188.399)(88,190.998)(89,193.292)(90,195.578)(91,197.872)(92,199.709)(93,202.222)(94,204.332)(95,206.663)(96,208.943)(97,211.144)(98,213.477)(99,215.874)(100,218.369) 
};

\addplot [
color=blue,
solid,
line width=2.0pt
]
coordinates{
 (1,2.66913)(2,5.02936)(3,7.42004)(4,9.58406)(5,11.8158)(6,13.7244)(7,15.3309)(8,17.0641)(9,18.6335)(10,20.0709)(11,21.5253)(12,22.9503)(13,24.311)(14,26.0458)(15,27.821)(16,29.5414)(17,30.9745)(18,32.845)(19,34.5082)(20,36.2892)(21,37.7321)(22,39.2279)(23,41.1306)(24,43.2068)(25,44.8386)(26,46.71)(27,48.5398)(28,50.2384)(29,51.7465)(30,53.4887)(31,55.4763)(32,57.3078)(33,58.9129)(34,60.6276)(35,62.4268)(36,64.0252)(37,66.1135)(38,67.9215)(39,70.081)(40,72.2261)(41,73.7824)(42,75.368)(43,77.5138)(44,79.7387)(45,81.8582)(46,83.8013)(47,85.9264)(48,87.8901)(49,90.0366)(50,92.2027)(51,94.1769)(52,96.2353)(53,98.3443)(54,100.125)(55,102.063)(56,104.163)(57,106.191)(58,108.227)(59,110.378)(60,112.494)(61,114.208)(62,116.38)(63,118.227)(64,120.239)(65,122.353)(66,124.527)(67,126.318)(68,128.316)(69,130.56)(70,132.899)(71,135.198)(72,137.053)(73,139.214)(74,141.138)(75,143.245)(76,145.568)(77,147.787)(78,150.036)(79,152.117)(80,154.276)(81,156.301)(82,158.105)(83,160.41)(84,162.178)(85,164.613)(86,166.646)(87,168.808)(88,171.067)(89,173.089)(90,175.196)(91,177.802)(92,179.99)(93,182.04)(94,184.147)(95,186.481)(96,188.469)(97,190.808)(98,192.949)(99,195.009)(100,197.654) 
};

\end{axis}
\end{tikzpicture}%
   \endpgfgraphicnamed%
 
  \caption{Regret, off-model tasks.}
  \label{fig:regret-off}
\end{figure}
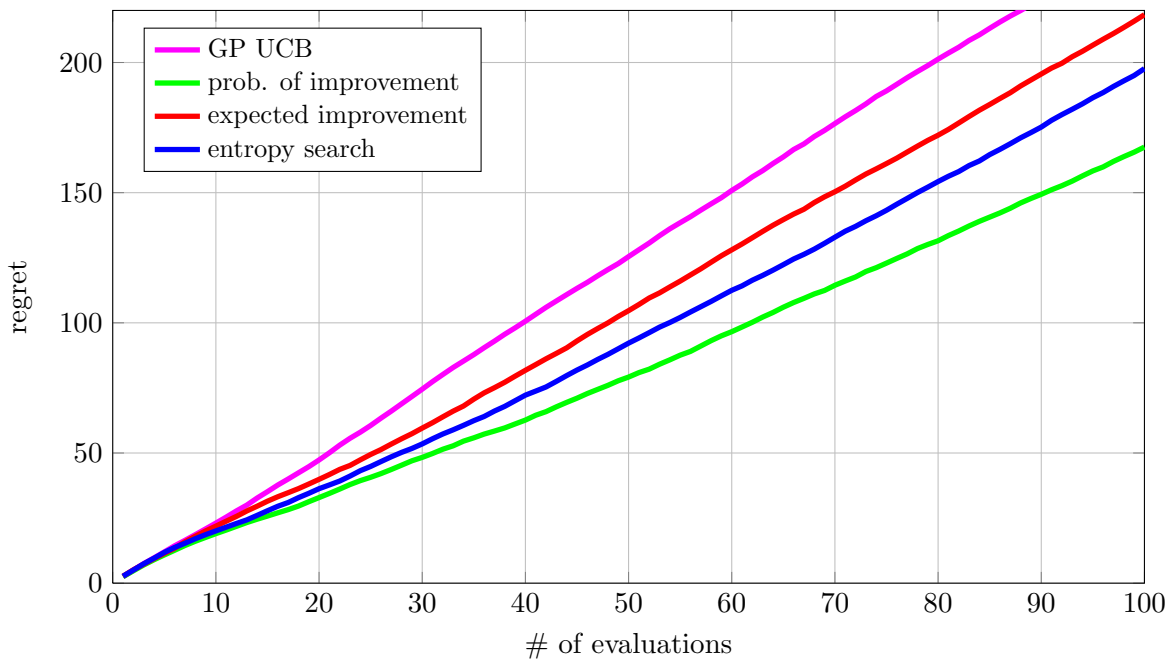

In the previous section, the algorithms attempted to find minima of
functions sampled from the prior used by the algorithms themselves. In
real applications, one can rarely hope to be so lucky, but
hierarchical inference can be used to generalize the prior and
construct a relatively general algorithm. But what if even the
hierarchically extended prior class does not contain the true
function? Qualitatively, it is clear that, beyond a certain point of
model-mismatch, all algorithms can be made to perform arbitrarily
badly. The poor performance of local optimizers (which may be
interpreted as building a quadratic model) in the previous section is
an example of this effect. In this section, we present results of the
same kind of experiments as in the previous section, but on a set of
30 two-dimensional functions sampled from a Gaussian process prior
with \emph{rational quadratic} kernel, with the same length scale and
signal variance as above, and scale mixture parameter $\alpha=1$ (see
Equation \eqref{eq:6}). This means samples evolve over an infinite
number of different length scales, including both longer and shorter
scales than those covered by the priors of the algorithms (Figure
\ref{fig:SE_RQ}). Figure \ref{fig:fval_offmodel} shows error on
function values, Figure \ref{fig:xval_offmodel} Euclidean error in
input space, Figure \ref{fig:regret-off} regret. Note the different
scales for the ordinate axes relative to the corresponding previous
plots: While Entropy Search still (barely) outperforms the
competitors, all three algorithms perform worse than before; and their
errors become more similar to each other. However, they still manage
to discover good regions in the domain, demonstrating a certain
robustness to model-mismatch.

\section{Conclusion}
\label{sec:conclusion}

This paper presented a new probabilistic paradigm for global
optimization, as an inference problem on the minimum of the function,
rather than the problem of collecting iteratively lower and lower
function values. We argue that this description is closer to
practitioners' requirements than classic response surface
optimization, bandit algorithms, or other, heuristic, global
optimization algorithms. In the main part of the paper, we constructed
Entropy Search, a practical probabilistic global optimization
algorithm, using a series of analytic assumptions and numerical
approximations: A particular family of priors over functions (Gaussian
processes); constructing the belief $\pmin$ over the location of the
minimum on an irregular grid to deal with the curse of dimensionality;
and using Expectation Propagation toward an efficient analytic
approximation. The Gaussian belief allows analytic probabilistic
predictions of the effect of future datapoints, from which we
constructed a first-order approximation of the expected change in
relative entropy of $\pmin$ to a base measure. For completeness, we
also pointed out some already known analytic properties of Gaussian
process measures that can be used to generalize this algorithm. We
showed that the resulting algorithm outperforms both directly and
distantly related competitors through its more elaborate,
probabilistic description of the problem. This increase in performance
is exchanged for somewhat increased computational cost (Entropy Search
costs are a constant multiple of that of classic Gaussian process
global optimizers). So this algorithm is more suited for problems were
evaluating the function itself carries considerable
cost. Nevertheless, it provides a natural description of the
optimization problem, by focusing on the performance under a loss
function at the horizon, rather than function values returned during
the optimization process. It allows the practitioner to explicitly
encode prior knowledge in a flexible way, and adapts its behavior to
the user's loss function.

\section*{Acknowledgments}
\label{sec:acknowledgments}

We would like to thank Martin Kiefel for valuable feedback, as well as
Tom Minka for an interesting discussion.

\appendix

\section{Mathematical Appendix}
\label{sec:math-append}

The notation in Equation \eqref{eq:1} can be read, sloppily, to mean
``$\pmin(x)$ is the probability that the value of $f$ at $x$ is lower
than at any other $\tilde{x}\in I$''. For a continuous domain, though,
there are uncountably many other $\tilde{x}$. To give more precise
meaning to this notation, consider the following argument. Let there
be a sequence of locations $\{x_i\}_{i=1,\dots,N}$, such that for
$N\to\infty$ the density of points at each location converges to a
measure $m(x)$ nonzero on every open neighborhood in $I$. If the
stochastic process $p(f)$ is sufficiently regular to ensure samples
are almost surely continuous (see footnote in Section
\ref{sec:gauss-proc-meas}), then almost every sample can be
approximated arbitrarily well by a staircase function with steps of
width $m(x_i)/N$ at the locations $x_i$, in the sense that $\forall
\epsilon > 0\; \exists N_0>0$ such that, $\forall N>N_0 :
|f(x)-f(\argmin_{x_j, j=1,\dots,N} |x-x_j|)| < \epsilon$, where
$|\cdot|$ is a norm (all norms on finite-dimensional vector spaces are
equivalent). This is the original reason why samples from sufficiently
regular Gaussian processes can be plotted using finitely many points,
in the way used in this paper. We now \emph{define} the notation used
in Equation \eqref{eq:1} to mean the following limit, where it exists.
\begin{multline}
  \label{eq:14}\raisetag{2cm}
  \pmin(x) = \int p(f) \prod_{\tilde{x} \neq x} \theta(f(\tilde{x})-f(x)) \d f \\
  \equiv \lim_{\stackrel{N \to \infty}{|x_i-x_{i-1}| \cdot N\to m(x)}}
  \int p[f(\{x_i\}_{i=1,\dots,N})] \prod_{i=1; i\neq j} ^N \theta [
  f(x_i) - f(x_j) ] \d f(\{x_i\}_{i=1,\dots,N}) \cdot |x_i-x_{i-1}|
  \cdot N
\end{multline}
In words: The ``infinite product'' is meant to be the limit of
finite-dimensional integrals with an increasing number of factors and
dimensions, where this limit exists. In doing so, we have side-stepped
the issue of whether this limit exists for any particular Gaussian
process (i.e.\ kernel function). We do so because the theory of
suprema of stochastic processes is highly nontrivial. We refer the
reader to a friendly but demanding introduction to the topic by
\citet{AdlerSuprema}. From our applied standpoint, the issue of
whether \eqref{eq:14} is well defined for a particular Gaussian prior
is secondary: If it is known that the true function is continuous and
bounded, than it has a well-defined supremum, and the prior should
reflect this knowledge by assigning sufficiently regular beliefs. If
the actual prior is such that we expect the function to be
discontinuous, it should be clear that optimization is extremely
challenging anyway. We conjecture that the finer details of the region
between these two domains have little relevance for communities
interested in optimization.

\bibliography{../bibfile}

\begin{thebibliography}{48}
\providecommand{\natexlab}[1]{#1}
\providecommand{\url}[1]{\texttt{#1}}
\expandafter\ifx\csname urlstyle\endcsname\relax
  \providecommand{\doi}[1]{doi: #1}\else
  \providecommand{\doi}{doi: \begingroup \urlstyle{rm}\Url}\fi

\bibitem[Adler(1981)]{adler1981geometry}
R.J. Adler.
\newblock \emph{The geometry of random fields}.
\newblock Wiley, 1981.

\bibitem[Adler(1990)]{AdlerSuprema}
R.J. Adler.
\newblock An introduction to continuity, extrema, and related topics for
  general {G}aussian processes.
\newblock \emph{Lecture Notes-Monograph Series}, 12:\penalty0
  i--iii+v--vii+ix+1--55, 1990.

\bibitem[Benoit(1924)]{Cholesky}
Benoit.
\newblock {N}ote s{\^u}re une m{\'e}thode de r{\'e}solution des {\'e}quations
  normales provenant de l'application de la m{\'e}thode des moindres carr{\'e}s
  a un syst{\`e}me d'{\'e}quations lin{\'e}aires en nombre inf{\'e}rieure a
  celui des inconnues. {A}pplication de la m{\'e}thode a la r{\'e}solution d'un
  syst{\`e}me d{\'e}fini d'{\'e}quations lin{\'e}aires. (proc{\'e}d{\'e} du
  {C}ommandant {C}holesky).
\newblock \emph{Bulletin geodesique}, 7\penalty0 (1):\penalty0 67--77, 1924.

\bibitem[Boyd and Vandenberghe(2004)]{boyd2004convex}
S.P. Boyd and L.~Vandenberghe.
\newblock \emph{Convex Optimization}.
\newblock Cambridge Univ Press, 2004.

\bibitem[Broyden et~al.(1965)]{broyden1965class}
C.G. Broyden et~al.
\newblock A class of methods for solving nonlinear simultaneous equations.
\newblock \emph{Math. Comp}, 19\penalty0 (92):\penalty0 577--593, 1965.

\bibitem[Byrd et~al.(1999)Byrd, Hribar, and Nocedal]{byrd1999interior}
R.H. Byrd, M.E. Hribar, and J.~Nocedal.
\newblock An interior point algorithm for large-scale nonlinear programming.
\newblock \emph{SIAM Journal on Optimization}, 9\penalty0 (4):\penalty0
  877--900, 1999.

\bibitem[Byrd et~al.(2000)Byrd, Gilbert, and Nocedal]{byrd2000trust}
R.H. Byrd, J.C. Gilbert, and J.~Nocedal.
\newblock A trust region method based on interior point techniques for
  nonlinear programming.
\newblock \emph{Mathematical Programming}, 89\penalty0 (1):\penalty0 149--185,
  2000.

\bibitem[Coleman and Li(1994)]{coleman1994convergence}
T.F. Coleman and Y.~Li.
\newblock On the convergence of interior-reflective newton methods for
  nonlinear minimization subject to bounds.
\newblock \emph{Mathematical programming}, 67\penalty0 (1):\penalty0 189--224,
  1994.

\bibitem[Coleman and Li(1996)]{coleman1996interior}
T.F. Coleman and Y.~Li.
\newblock An interior trust region approach for nonlinear minimization subject
  to bounds.
\newblock \emph{SIAM Journal on Optimization}, 6\penalty0 (2):\penalty0
  418--445, 1996.

\bibitem[Cox(1946)]{cox1946probability}
R.T. Cox.
\newblock {Probability, frequency and reasonable expectation}.
\newblock \emph{American Journal of Physics}, 14\penalty0 (1):\penalty0 1--13,
  1946.

\bibitem[Cunningham et~al.(2011)Cunningham, Hennig, and Lacoste-Julien]{EPMGP}
J.~Cunningham, P.~Hennig, and S.~Lacoste-Julien.
\newblock Gaussian probabilities and expectation propagation.
\newblock \emph{under review. Preprint at arXiv:1111.6832 [stat.ML]}, November
  2011.

\bibitem[Fletcher(1970)]{fletcher1970new}
R.~Fletcher.
\newblock A new approach to variable metric algorithms.
\newblock \emph{The Computer Journal}, 13\penalty0 (3):\penalty0 317, 1970.

\bibitem[Goldfarb(1970)]{goldfarb1970family}
D.~Goldfarb.
\newblock A family of variable metric updates derived by variational means.
\newblock \emph{Mathematics of Computing}, 24\penalty0 (109):\penalty0 23--26,
  1970.

\bibitem[Gr{\"u}new{\"a}lder et~al.(2010)Gr{\"u}new{\"a}lder, Audibert, Opper,
  and Shawe-Taylor]{grunewalder2010regret}
S.~Gr{\"u}new{\"a}lder, J.Y. Audibert, M.~Opper, and J.~Shawe-Taylor.
\newblock Regret bounds for {G}aussian process bandit problems.
\newblock In \emph{Proceedings of the 14th International Conference on
  Artificial Intelligence and Statistics (AISTATS)}, 2010.

\bibitem[Hansen and Ostermeier(2001)]{hansen2001completely}
N.~Hansen and A.~Ostermeier.
\newblock Completely derandomized self-adaptation in evolution strategies.
\newblock \emph{Evolutionary computation}, 9\penalty0 (2):\penalty0 159--195,
  2001.

\bibitem[Hennig(2011)]{GaussianRL}
P.~Hennig.
\newblock Optimal reinforcement learning for {G}aussian systems.
\newblock In \emph{Advances in Neural Information Processing Systems}, 2011.

\bibitem[Hestenes and Stiefel(1952)]{hestenes1952methods}
M.R. Hestenes and E.~Stiefel.
\newblock {Methods of conjugate gradients for solving linear systems}.
\newblock \emph{Journal of Research of the National Bureau of Standards},
  49\penalty0 (6):\penalty0 409--436, 1952.

\bibitem[It{\=o}(1951)]{ito1951stochastic}
K.~It{\=o}.
\newblock On stochastic differential equations.
\newblock \emph{Memoirs of the American Mathematical Society}, 4, 1951.

\bibitem[Jaynes and Bretthorst(2003)]{jaynes2003probability}
E.T. Jaynes and G.L. Bretthorst.
\newblock \emph{{Probability Theory: the Logic of Science}}.
\newblock Cambridge University Press, 2003.

\bibitem[Jones et~al.(1998)Jones, Schonlau, and Welch]{jones1998efficient}
D.R. Jones, M.~Schonlau, and W.J. Welch.
\newblock Efficient global optimization of expensive black-box functions.
\newblock \emph{Journal of Global optimization}, 13\penalty0 (4):\penalty0
  455--492, 1998.

\bibitem[Kleinberg(2005)]{kleinberg2005nearly}
R.~Kleinberg.
\newblock Nearly tight bounds for the continuum-armed bandit problem.
\newblock \emph{Advances in Neural Information Processing Systems}, 18, 2005.

\bibitem[Kolmogorov(1933)]{kolmogorov_axioms}
A.N. Kolmogorov.
\newblock Grundbegriffe der {W}ahrscheinlichkeitsrechnung.
\newblock \emph{Ergebnisse der Mathematik und ihrer Grenzgebiete}, 2, 1933.

\bibitem[Krige(1951)]{krigestatistical}
D.G. Krige.
\newblock A statistical approach to some basic mine valuation and allied
  problems at the {W}itwatersrand.
\newblock Master's thesis, University of Witwatersrand, 1951.

\bibitem[Kullback and Leibler(1951)]{kullback1951information}
S.~Kullback and R.A. Leibler.
\newblock {On information and sufficiency}.
\newblock \emph{Annals of Mathematical Statistics}, 22\penalty0 (1):\penalty0
  79--86, 1951.

\bibitem[Lazard-Holly and Holly(2003)]{Holly_NInt}
H.~Lazard-Holly and A.~Holly.
\newblock Computation of the probability that a $d$-dimensional normal variable
  belongs to a polyhedral cone with arbitrary vertex.
\newblock Technical report, Mimeo, 2003.

\bibitem[Lizotte(2008)]{lizotte2008practical}
D.J. Lizotte.
\newblock \emph{Practical {B}ayesian Optimization}.
\newblock PhD thesis, University of Alberta, 2008.

\bibitem[MacKay(1998{\natexlab{a}})]{mackay1998choice}
D.J.C. MacKay.
\newblock {Choice of basis for Laplace approximation}.
\newblock \emph{Machine Learning}, 33\penalty0 (1):\penalty0 77--86,
  1998{\natexlab{a}}.

\bibitem[MacKay(1998{\natexlab{b}})]{mackay1998introduction}
D.J.C. MacKay.
\newblock Introduction to {G}aussian processes.
\newblock \emph{NATO ASI Series F Computer and Systems Sciences}, 168:\penalty0
  133--166, 1998{\natexlab{b}}.

\bibitem[Mat{\'e}rn(1960)]{matérn1960spatial}
B.~Mat{\'e}rn.
\newblock Spatial variation.
\newblock \emph{Meddelanden fran statens Skogsforskningsinstitut}, 49\penalty0
  (5), 1960.

\bibitem[Minka(2000)]{minka2000deriving}
T.P. Minka.
\newblock Deriving quadrature rules from {G}aussian processes.
\newblock Technical report, Statistics Department, Carnegie Mellon University,
  2000.

\bibitem[Minka(2001)]{EP_Minka}
T.P. Minka.
\newblock Expectation {P}ropagation for approximate {B}ayesian inference.
\newblock In \emph{Proceedings of the 17th Conference in Uncertainty in
  Artificial Intelligence}, pages 362--369, San Francisco, CA, USA, 2001.
  Morgan Kaufmann.
\newblock ISBN 1-55860-800-1.

\bibitem[Murray and Adams(2010)]{murray2010slice}
I.~Murray and R.P. Adams.
\newblock {Slice sampling covariance hyperparameters of latent Gaussian
  models}.
\newblock \emph{arXiv:1006.0868}, 2010.

\bibitem[Nocedal and Wright(1999)]{nocedal1999numerical}
J.~Nocedal and S.J. Wright.
\newblock \emph{Numerical optimization}.
\newblock Springer Verlag, 1999.

\bibitem[Osborne et~al.(2009)Osborne, Garnett, and
  Roberts]{osborne2009gaussian}
M.A. Osborne, R.~Garnett, and S.J. Roberts.
\newblock {G}aussian processes for global optimization.
\newblock In \emph{3rd International Conference on Learning and Intelligent
  Optimization (LION3)}, 2009.

\bibitem[Plackett(1954)]{plackett1954reduction}
R.L. Plackett.
\newblock {A reduction formula for normal multivariate integrals}.
\newblock \emph{Biometrika}, 41\penalty0 (3-4):\penalty0 351, 1954.

\bibitem[Powell(1978{\natexlab{a}})]{powell1978convergence}
M.J.D. Powell.
\newblock The convergence of variable metric methods for nonlinearly
  constrained optimization calculations.
\newblock \emph{Nonlinear programming}, 3\penalty0 (0):\penalty0 27--63,
  1978{\natexlab{a}}.

\bibitem[Powell(1978{\natexlab{b}})]{powell1978fast}
M.J.D. Powell.
\newblock A fast algorithm for nonlinearly constrained optimization
  calculations.
\newblock \emph{Numerical analysis}, pages 144--157, 1978{\natexlab{b}}.

\bibitem[Rasmussen and Williams(2006)]{RasmussenWilliams}
C.E. Rasmussen and C.K.I. Williams.
\newblock \emph{Gaussian Processes for Machine Learning}.
\newblock MIT Press, 2006.

\bibitem[Schmitt(2004)]{schmitt2004theory}
L.M. Schmitt.
\newblock Theory of genetic algorithms ii: models for genetic operators over
  the string-tensor representation of populations and convergence to global
  optima for arbitrary fitness function under scaling.
\newblock \emph{Theoretical Computer Science}, 310\penalty0 (1-3):\penalty0
  181--231, 2004.

\bibitem[Seeger(2008)]{SeegerEP}
M.~Seeger.
\newblock Expectation propagation for exponential families.
\newblock Technical report, U.C. Berkeley, 2008.

\bibitem[Shanno(1970)]{shanno1970conditioning}
D.F. Shanno.
\newblock Conditioning of quasi-{N}ewton methods for function minimization.
\newblock \emph{Mathematics of computation}, 24\penalty0 (111):\penalty0
  647--656, 1970.

\bibitem[Shannon(1948)]{shannon1948mathematical}
C.E. Shannon.
\newblock A mathematical theory of communication.
\newblock \emph{Bell System Technical Journal}, 27, 1948.

\bibitem[Srinivas et~al.(2010)Srinivas, Krause, Kakade, and
  Seeger]{srinivasgaussian}
N.~Srinivas, A.~Krause, S.~Kakade, and M.~Seeger.
\newblock {G}aussian process optimization in the bandit setting: No regret and
  experimental design.
\newblock In \emph{International Conference on Machine Learning}, 2010.

\bibitem[Thompson and Neal(2010)]{thompson2010slice}
M.B. Thompson and R.M. Neal.
\newblock Slice sampling with adaptive multivariate steps: The shrinking-rank
  method.
\newblock \emph{Arxiv preprint arXiv:1011.4722}, 2010.

\bibitem[Uhlenbeck and Ornstein(1930)]{uhlenbeck1930theory}
G.E. Uhlenbeck and L.S. Ornstein.
\newblock On the theory of the {B}rownian motion.
\newblock \emph{Physical Review}, 36\penalty0 (5):\penalty0 823, 1930.

\bibitem[van~der Vaart and van Zanten(2011)]{van2011information}
A.W. van~der Vaart and J.H. van Zanten.
\newblock Information rates of nonparametric {G}aussian process methods.
\newblock \emph{Journal of Machine Learning Research}, 12:\penalty0 2095--2119,
  2011.

\bibitem[Waltz et~al.(2006)Waltz, Morales, Nocedal, and
  Orban]{waltz2006interior}
R.A. Waltz, J.L. Morales, J.~Nocedal, and D.~Orban.
\newblock An interior algorithm for nonlinear optimization that combines line
  search and trust region steps.
\newblock \emph{Mathematical Programming}, 107\penalty0 (3):\penalty0 391--408,
  2006.

\bibitem[Wiener and Masani(1957)]{wiener1957prediction}
N.~Wiener and P.~Masani.
\newblock The prediction theory of multivariate stochastic processes.
\newblock \emph{Acta Mathematica}, 98\penalty0 (1):\penalty0 111--150, 1957.

\end{thebibliography}

\end{document}